%% file: main.tex
\documentclass[letterpaper,twocolumn,10pt]{article}
\usepackage{usenix}
\usepackage{graphicx}
\usepackage{url}
\usepackage{xcolor}
\usepackage{multirow}
\usepackage[inline]{enumitem}

\newif\ifsubmit\submitfalse

\usepackage{booktabs}
\usepackage{tabularx}
\usepackage{adjustbox}
\usepackage{booktabs,adjustbox,array,makecell}
\usepackage{xcolor}

\usepackage{tikz}
\usepackage[most]{tcolorbox}
\usepackage{pgf-pie}
\usepackage{tikz}
\usetikzlibrary{calc}
\usepackage{xparse}
\usepackage{ifthen}
\usetikzlibrary{arrows.meta,positioning,fit,calc,shapes.geometric}
\usepackage{xspace}
\usepackage{amssymb}

\usepackage[Tol]{colorblind}

\colorlet{pieGrey}{T-Q-B0} 
\colorlet{textGrey}{T-S-DR0}
\colorlet{pieGreenD}{T-D-PG8}     
\colorlet{pieGreenL}{T-D-PG7}
\colorlet{textGreenD}{T-D-PG9}
\colorlet{textGreenL}{T-D-PG8}

\colorlet{pieBlueD}{T-D-BR2}   
\colorlet{pieBlueL}{T-D-BR3}
\colorlet{textBlueD}{T-D-BR1}   
\colorlet{textBlueL}{T-D-BR2}

\colorlet{piePurpleD}{T-D-PG2}     
\colorlet{piePurpleL}{T-D-PG3}
\colorlet{textPurpleD}{T-D-PG1}     
\colorlet{textPurpleL}{T-D-PG2}

\colorlet{pieRedD}{T-D-BR8}    
\colorlet{pieRedL}{T-D-BR7}
\colorlet{textRedD}{T-D-BR9}    
\colorlet{textRedL}{T-D-BR8}

\usepackage{acronym}

  \acrodef{DRL}{Deep Reinforcement Learning}
  \acrodef{RL}{Reinforcement Learning}
  \acrodef{ACD}{Autonomous Cyber defense}
  \acrodef{AML}{Adversarial Machine Learning}
  \acrodef{WST}{Web Security Testing}
  \acrodef{MDP}[MDP]{Markov Decision Process}
  \acrodefplural{MDP}[MDPs]{Markov Decision Processes}
  \acrodef{POMDP}[POMDP]{Partially Observable \ac{MDP}}
  \acrodefplural{POMDP}[POMDPs]{Partially Observable \acp{MDP}}

\newcommand{\mc}{\hyperref[sec:mc]{M-MC}\xspace}
\newcommand{\mcfull}{Modeling Correctness\xspace}
\newcommand{\ms}{\hyperref[sec:ms]{M-MS}\xspace}
\newcommand{\msfull}{\ac{MDP} Specification\xspace}
\newcommand{\po}{\hyperref[sec:po]{M-PO}\xspace}
\newcommand{\pofull}{Partial Observability\xspace}
\newcommand{\hr}{\hyperref[sec:hr]{T-HR}\xspace}
\newcommand{\hrfull}{Hyperparameter Reporting\xspace}
\newcommand{\va}{\hyperref[sec:va]{T-VA}\xspace}
\newcommand{\vafull}{Variance Analysis\xspace}
\newcommand{\pc}{\hyperref[sec:pc]{T-PC}\xspace}
\newcommand{\pcfull}{Policy Convergence\xspace}
\newcommand{\am}{\hyperref[sec:am]{E-AM}\xspace}
\newcommand{\amfull}{Application Motivation\xspace}
\newcommand{\ga}{\hyperref[sec:ga]{E-GA}\xspace}
\newcommand{\gafull}{Gain Attribution\xspace}
\newcommand{\ec}{\hyperref[sec:ec]{E-EC}\xspace}
\newcommand{\ecfull}{Environment Complexity\xspace}
\newcommand{\ua}{\hyperref[sec:ua]{D-UA}\xspace}
\newcommand{\uafull}{Underlying Assumptions\xspace}
\newcommand{\ns}{\hyperref[sec:ns]{D-NS}\xspace}
\newcommand{\nsfull}{Non-Stationarity\xspace}
\newcommand{\etal}{{\it et al.\xspace}}%
\newcommand\code[1]{{\small\texttt{#1}}}
\newcommand{\eg}{{\it e.g.,}\xspace}
\newcommand{\ie}{{\it i.e.,}\xspace}
\newcommand{\drlsec}{{\sc Drl4Sec}\xspace}
\newcommand\sqirl{\textsc{Sqirl}}

\newtcolorbox{boxK}{
    sharpish corners, 
    boxrule = 0pt,
    toprule = 4.5pt, 
    enhanced,
    fuzzy shadow = {0pt}{-2pt}{-0.5pt}{0.5pt}{black!35} 
}
\newtcolorbox{boxME}{
    sharpish corners,
    boxrule = 0pt,
    toprule = 4.5pt, 
    enhanced,
    colframe=pieGreenD,
    fuzzy shadow = {0pt}{-2pt}{-0.5pt}{0.5pt}{black!35} 
}
\newtcolorbox{boxTA}{
    sharpish corners, 
    boxrule = 0pt,
    toprule = 4.5pt, 
    enhanced,
    colframe=pieBlueD,
    fuzzy shadow = {0pt}{-2pt}{-0.5pt}{0.5pt}{black!35} 
}
\newtcolorbox{boxEA}{
    sharpish corners, 
    boxrule = 0pt,
    toprule = 4.5pt, 
    enhanced,
    colframe=piePurpleD,
    fuzzy shadow = {0pt}{-2pt}{-0.5pt}{0.5pt}{black!35} 
}
\newtcolorbox{boxDA}{
    sharpish corners,
    boxrule = 0pt,
    toprule = 4.5pt, 
    enhanced,
    colframe=pieRedD,
    fuzzy shadow = {0pt}{-2pt}{-0.5pt}{0.5pt}{black!35} 
}

\title{\Large \bf SoK: The Pitfalls of Deep Reinforcement Learning for Cybersecurity}

\author{
    {\rm  
        Shae McFadden\textsuperscript{\rm$\dagger$\rm$\ddagger$\rm$\mathsection$}, 
        Myles Foley\textsuperscript{\rm$\ddagger$\rm*}, 
        Elizabeth Bates\textsuperscript{\rm$\ddagger$}, 
        Ilias Tsingenopoulos\textsuperscript{\rm$\P$},
    } \\ {\rm 
        Sanyam Vyas\textsuperscript{\rm$\parallel$\rm$\ddagger$},
        Vasilios Mavroudis\textsuperscript{\rm$\dagger$}, 
        Chris Hicks\textsuperscript{\rm$\ddagger$}, 
        Fabio Pierazzi\textsuperscript{\rm$\mathsection$}
    } \\ 
        \textsuperscript{\rm  $\dagger$}King's College London, 
        \textsuperscript{\rm $\ddagger$}The Alan Turing Institute, 
        \textsuperscript{\rm  $\mathsection$}University College London, 
    \\ 
        \textsuperscript{\rm  $\parallel$}Cardiff University, 
        \textsuperscript{\rm  $\P$}KU Leuven,
        \textsuperscript{\rm  *}Devotion AI Labs
} 

\begin{document}

\maketitle

\begin{abstract}
    \ac{DRL} has achieved remarkable success in domains requiring sequential decision-making, motivating its application to cybersecurity problems. However, transitioning \ac{DRL} from laboratory simulations to bespoke cyber environments can introduce numerous issues. This is further exacerbated by the often adversarial, non-stationary, and partially-observable nature of most cybersecurity tasks. In this paper, we identify and systematize 11 methodological pitfalls that frequently occur in DRL for cybersecurity (\drlsec) literature across the stages of environment modeling, agent training, performance evaluation, and system deployment. By analyzing $66$ significant \drlsec papers (2018-2025), we quantify the prevalence of each pitfall and find an average of over five pitfalls per paper. We demonstrate the practical impact of these pitfalls using controlled experiments in (i) autonomous cyber defense, (ii) adversarial malware creation, and (iii) web security testing environments. Finally, we provide actionable recommendations for each pitfall to support the development of more rigorous and deployable DRL-based security systems.
\end{abstract}

\section{Introduction}

\acl{DRL} (DRL) has successfully improved the speed of matrix multiplication~\cite{fawzi_discovering_2022}; discovered new protein structures~\cite{jumper_highly_2021}; and, achieved superhuman performance at game playing~\cite{mnih2013playing,wurman_outracing_2022}.
Through continuous interaction with a DRL environment (often simply referred to as the \emph{environment}), \ac{DRL} agents learn sequential decision-making policies by optimizing the expected cumulative reward over time~\cite{sutton2018reinforcement}. Unsurprisingly, these capabilities have motivated applications in cybersecurity, where systems must make adaptive decisions in dynamic, adversarial environments. DRL applications range from detecting malicious activity~\cite{tong2020finding, tharewal2022intrusion, praveena2022optimal, McFadden2026DRMD} to evading detection~\cite{anderson2018learning, wu2019evading, zhao2021structural, wang2021crafting, tsingenopoulos2022captcha, rigaki2023power, zhan2023psp, tsingenopoulos2024train, chen2024llm} and from autonomously defending networks~\cite{han2018reinforcement, kvasov2023simulating, bates2023reward, goel2024optimizing, terranova2024leveraging, foleyCAGEI22, foleyCAGEII22, hicksCAGE3_23} to discovering vulnerabilities~\cite{li2022alphaprog, foley2025apirl, lee2022link, erdHodi2021simulating, al2023sqirl, corradini2024deeprest, eom2024fuzzing, faillon2024better, mcfadden2024wendigo}.

However, the transition to cybersecurity introduces fundamental challenges compared to traditional domains where \ac{DRL} has previously excelled, such as game playing~\cite{mnih2013playing}, robotics~\cite{dambrosio_achieving_2024}, and computer chip design~\cite{mirhoseini_graph_2021}. In cybersecurity, adversaries actively adapt their strategies to evade detection, systems evolve over time, and defenders typically observe limited information about attacker intentions and system state~\cite{moore_fundamental_2025}. These characteristics violate core assumptions of \acp{MDP}, the model for DRL learning tasks, that presume fixed transition dynamics and complete state observability~\cite{sutton2018reinforcement}, which can cause agents to be vulnerable to adversaries~\cite{simon_learning_2025}. In reality, most security tasks are better characterized as \acp{POMDP}, where agents must infer hidden dynamics from incomplete knowledge: an \textit{observation} of the system state. Further compounding these challenges are additional task-specific requirements, from information constraints (privacy and access limitations) to practical deployment capabilities (query budgets, real-time constraints). Failing to address these can produce agents that perform well in controlled simulations, however, fail unpredictably and dangerously when deployed in operational environments~\cite{skalse_defining_2022,nikishin_primacy_2022}.

\begin{figure*}[t]
    \centering
    \includegraphics[width=\linewidth]{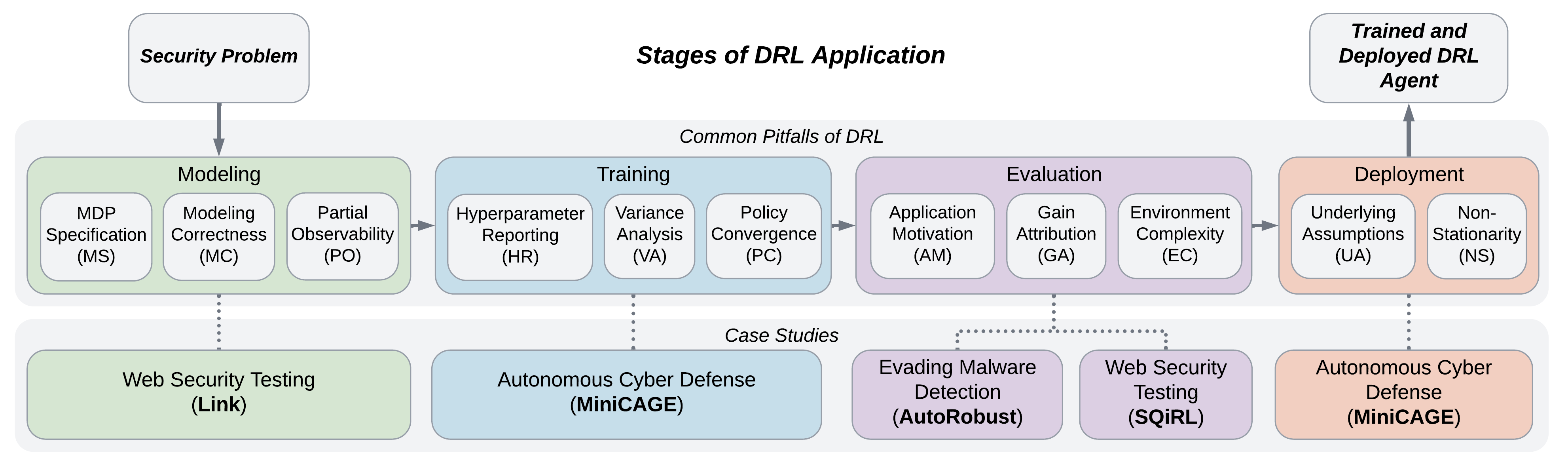}
    \caption{Common pitfalls of \ac{DRL} applied to cybersecurity, organized by development stage and separate case studies.}
    \label{fig:drl-pitfalls-workflow}
\end{figure*}

In this paper, we systematize 11 recurring methodological pitfalls that emerge when \ac{DRL} is employed in cybersecurity (\drlsec). As shown in Figure~\ref{fig:drl-pitfalls-workflow}, we organize these pitfalls across the four stages of agent development: environment modeling (\textcolor{pieGreenD}{green}), agent training (\textcolor{pieBlueD}{blue}), performance evaluation (\textcolor{piePurpleD}{purple}), and system deployment (\textcolor{pieRedD}{red}). Through systematic analysis of 66 \drlsec papers, published between 2018 and 2025, we quantify the prevalence and severity of each pitfall. Our review reveals widespread issues where \emph{every paper contains at least two pitfalls}, with an average of 5.8.
Furthermore, across the individual pitfalls, 28.8\% of papers fall victim to the least common pitfall (\uafull), with the most common being present in 71.2\% of papers (\pcfull).
Through experiments across three \drlsec domains, we illustrate how each of the pitfalls can cause performance degradation or produce misleading results. Figure~\ref{fig:drl-pitfalls-workflow} shows the pitfalls and specific case study domains used for each section in this paper.

Collectively, these pitfalls obscure whether reported improvements stem from genuine algorithmic advances or rather from artifacts of simplified environments and incomplete evaluations. More critically, they create a dangerous gap between simulated and operational reality, leading to systems that provide a false sense of security when deployed. Our goal is not to discourage \ac{DRL} applications in cybersecurity, as it holds genuine promise, but rather to establish methodological standards that enable the community to distinguish robust, deployable solutions from those that succeed only in idealized settings.
In summary, we make the following contributions:
\begin{itemize}[leftmargin=*]

    \item \textbf{Prevalence Analysis} We systematically examine 66 \drlsec papers published between 2018 and 2025, quantifying the occurrence and severity of pitfalls across diverse security domains.
    
    \item \textbf{Pitfall Taxonomy} We define 11 pitfalls when adapting \ac{DRL} to cybersecurity, organized across four stages of development (modeling, training, evaluation, and deployment).
    
    \item \textbf{Impact Demonstration} Through experiments using existing \drlsec environments in three representative domains, we provide empirical evidence of the impact caused by these pitfalls.
    
    \item \textbf{Actionable Recommendations} We provide concrete guidance to mitigate each pitfall, supporting the development of more rigorous and reliable \drlsec research.
\end{itemize}

\begin{boxK}
    \textbf{Remark} This work identifies recurring pitfalls in \drlsec literature. These are not failures of \emph{individual} researchers but \emph{systematic} challenges that emerge when adapting \ac{DRL} to cybersecurity.
\end{boxK}

\section{Foundational Concepts} \label{sec:background}
    At its core, \ac{RL} problems learn optimal policies through environmental interaction formalized as MDPs. MDPs are defined by the tuple $(\mathcal{S}, \mathcal{A}, \mathcal{R}, \mathcal{T}, \gamma)$, consisting of: a state space $\mathcal{S}$, action space $\mathcal{A}$, reward function $\mathcal{R}: \mathcal{S} \times \mathcal{A} \rightarrow \mathbb{R}$, transition function $\mathcal{T}: \mathcal{S} \times \mathcal{A} \rightarrow \Delta(\mathcal{S})$ (where $\Delta(\mathcal{S})$ denotes probability distributions over states), and discount factor $\gamma \in [0,1]$~\cite{kaelbling1996reinforcement, arulkumaran2017deep, sutton2018reinforcement}. 
    An RL `agent' observes a state $s_t$, selects an action $a_t$ according to the policy $\pi$, then receives a reward $r_t$, and transitions to the next state $s_{t+1}$. The agent then learns a policy that maximizes expected cumulative discounted reward: $\mathbb{E}[\sum_{t=0}^{\infty} \gamma\ ^t r_t]$.
    Traditional RL algorithms such as Q-Learning~\cite{watkins1992q} maintain explicit tables of state-action values that do not generalize to unseen states, and become intractable for large state or action spaces, which is critical for cybersecurity applications~\cite{nguyen2021deep}. \ac{DRL} addresses this limitation by using deep neural networks as function approximators, allowing generalization to unseen states and scalability to high-dimensional problems~\cite{mnih2013playing}.
    The three primary paradigms of \ac{DRL} are: \textit{value-based methods}, which approximate the state-action values of Q-learning~\cite{mnih2013playing,van2016double,wang2016dueling}; \textit{policy-based methods}, which directly approximate the action distribution of the policy~\cite{schulman2015trust}; and, \textit{actor-critic methods} that combine both paradigms~\cite{mnih2016asynchronous,schulman2017proximal}.
    The \ac{MDP} formulation relies on the \textit{Markov property}: that the future state depends \emph{only} on the current state and action, not on the history of past states~\cite{chung1967markov}. This assumption is critical for \ac{DRL} but can be violated in cybersecurity settings where attackers maintain a hidden state (\eg compromised credentials, established persistence) not observable to defenders. In such cases, the problem is more accurately modeled as a \ac{POMDP}, where the agent receives observations $o_t$ from an observation function $\Omega: \mathcal{S} \rightarrow \mathcal{O}$ rather than directly receiving states~\cite{boutilier1999decision}. \acp{POMDP} require agents to maintain belief states or use recurrent architectures to infer the hidden state from observation histories.

\section{Review Methodology} \label{sec:review-method}

    We followed a similar paper collection and review framework as was used in \cite{DosDonts,evertz2025chasing}, however, as \drlsec is still a growing community we extend our search beyond just top-tier conferences (\eg USENIX Security or IEEE S\&P). Therefore, we conducted a systematic literature collection using IEEE Xplore, ACM DL, and Google Scholar. The search keywords can be found in Appendix~\ref{app:keywords}. 
    
    \textbf{Collection} We performed our collection in two rounds, first in April 2025 using only Google Scholar for initial pitfall identification and second in December 2025 using all search tools to complete the corpus and update citation counts. After the final collection, we performed de-duplication, resulting in 1,603 papers. Then, we performed the \emph{Significance Criteria} filtering to arrive at 363 papers. Finally, we screened the papers for \emph{Topic Criteria} filtering alongside requirements for language (English) and length ($>4$ pages) to produce the final 66 papers used in our work, with 37 from the April and 29 from the December collection.

    \textbf{Significance Criteria} To ensure academic impact and quality, papers must satisfy at least one of the following criteria: (1) published in a first- or second-tier (CORE-ranked\footnote{\url{https://portal.core.edu.au/conf-ranks/}} A* or A) venue in machine learning, software engineering, or security; or (2) demonstrate significant research impact by averaging more than 10 citations per year since publication from a reputable publisher (see Appendix~\ref{app:keywords}).

    \textbf{Topic Criteria} We define the scope of papers to be included in our work as: \textit{direct} applications of \ac{DRL} to \emph{cybersecurity} problems and have \ac{DRL} as \emph{the primary methodological contribution}. We then excluded several categories of related but out-of-scope work: (1) secondary studies (surveys, systematic reviews, \textit{etc.}); (2) research on the security \textit{of} DRL systems, \ie non-DRL-based attacks against DRL agents (which are not applied to \emph{cybersecurity} problems); (3) pure game-theoretic approaches without DRL; (4) non-deep RL methods (\eg tabular Q-learning, which fail in complex environments as discussed in Section~\ref{sec:background}); and (5) Multi-Agent RL (MARL) systems, a separate research domain (which require distinct formulations and analysis). 

    \textbf{Corpus Characteristics} The 66 papers in our corpus were published between 2018 and 2025 across a diverse set of cybersecurity tasks: \textit{adversarial sample creation} (15 papers, 22.7\%)~\cite{anderson2018learning, yang2020patchattack, wu2019evading, zhao2021structural, wang2021crafting, tsingenopoulos2022captcha, rigaki2023power, zhan2023psp, tsingenopoulos2024train, gohil2024attackgnn, chen2024llm, hore2025deep, apruzzese2020deep, randhawa2024deep, fang2019evading}; \textit{intrusion detection} (15 papers, 22.7\%)~\cite{tong2020finding, tharewal2022intrusion, praveena2022optimal, lopez2020application, caminero2019adversarial, mohamed2023deep, he2024reinforcement, yu2023deep, hsu2020deep, sethi2020deep, sharma2024batch, phan2022deepair, dos2022reinforcement, nie2021intrusion, li2023soft}; \textit{fuzzing and web security} (10 papers, 15.2\%)~\cite{bottinger2018deep, li2022alphaprog, romdhana2022deep, foley2025apirl, lee2022link, erdHodi2021simulating, al2023sqirl, corradini2024deeprest, eom2024fuzzing, faillon2024better}; \textit{autonomous cyber defense} (7 papers, 10.6\%)~\cite{han2018reinforcement, kvasov2023simulating, bates2023reward, zhang2023moving, goel2024optimizing, kim2022divergence, terranova2024leveraging}; \textit{penetration testing} (6 papers, 9.1\%)~\cite{yang2025behaviour, hu2020automated, chen2023gail, maeda2021automating, gangupantulu2022using, li2023innes}; \textit{cyber-physical systems security} (4 papers, 6.1\%)~\cite{wang2020coordinated, landen2022dragon, paudel2023reinforcement, le2024exploration}; \textit{malware detection} (4 papers, 6.1\%)~\cite{al2023malbot,basnet2025advanced,chatterjee2019detecting,deng2024ransomware}; \textit{blockchain security} (3 papers, 4.5\%)~\cite{hou2019squirrl, de2024guideenricher, su2022effectively}; and \textit{hardware security} (2 papers, 3.0\%)~\cite{gohil2022attrition, luo2023autocat}. These domains represent the major application areas of \drlsec, showing concentration in areas requiring complex sequential decision-making.

    \textbf{Review Process} We began with a pilot review of the 37 papers from the April collection to identify themes and commonalities. Using this review, we isolated the reoccurring issues in \drlsec papers and formalized the 11 pitfalls discussed in this work. 
    We then evaluated the complete 66 papers for each of the identified pitfalls. Each paper was independently evaluated by \textit{two reviewers} using the pitfall definitions provided in the following Sections~\ref{sec:pitfall-modeling}--\ref{sec:pitfall-deployment}. Reviewers classified each of the 11 pitfalls as \textit{present} (clearly and completely exhibited), \textit{partially present} (partial manifestations or acknowledged but unaddressed), or \textit{not present} (adequately handled or not applicable).
    Following the independent reviews, the initial prevalence scores achieved substantial agreement, according to a Cohen's $\kappa$ of 0.712 and the Landis–Koch interpretation~\cite{landis1977measurement}. With a 81.5\% total agreement on pitfall prevalence, the remaining disagreements (18.5\%) around the severity of pitfalls for the papers reviewed were resolved through discussion between the reviewers. 
    When evidence was ambiguous, we gave the benefit of the doubt to the paper by using the less severe pitfall classification to resolve the disagreement (\eg marking as \textit{not present} if reviews were split between \textit{not present} and \textit{partially present}).
    The final prevalence scores determined through this process form the basis of our analysis. Further breakdown of agreement, kappa, and interpretation can be found in Appendix~\ref{app:keywords}.
    
    For each pitfall, we represent its prevalence using a pie chart alongside its definition with present, partially present, and not present in dark, light, and gray colors, respectively. 

\section{Case Study Environments} \label{sec:case-study-envs} 
We conducted controlled experiments across three \drlsec domains to demonstrate the practical impact of the identified pitfalls. First, we consider \ac{ACD} as these environments are well established and often reused across subsequent works~\cite{foleyCAGEI22,foleyCAGEII22,applebaum_bridging_2022,bates2023reward,collyer_acd-g_2022,hicksCAGE3_23,emerson_cyborg_2024,bates2025less}. Then, we consider adversarial malware generation and web security testing as they are some of the most prominent applications of \drlsec across the 66 papers studied (22.7\% and 15.2\%, respectively, see Section~\ref{sec:review-method}). In this section, we provide a high level overview to the environments used in our case studies; while specific experiment details are provided in their respective sections. For all environments, their full \ac{MDP} specifications are included in Appendix~\ref{app:case_studies}.

    \textbf{Autonomous Cyber Defense} 
    For our experiments in \ac{ACD}, we use the MiniCAGE~\cite{emerson_cyborg_2024} environment, which replicates and extends the network defense task proposed in the CAGE 2 challenge~\cite{cage_cyborg_2023}, used in a number of subsequent works~\cite{foleyCAGEI22,bates2023reward,bates2025less,applebaum_bridging_2022}. The enterprise network to be defended comprises 13 hosts across three subnets: a user subnet (5 hosts), an enterprise subnet (3 hosts plus 1 defender host), and an operational subnet (4 hosts including a critical server). The defensive (blue) agent must prevent an attacker (red) from reaching the critical server. MiniCAGE contains two attacker strategies: the B-Line attacker proceeds directly through the network to compromise the critical server as quickly as possible, while the Meander attacker attempts to compromise the entire network while en route to the critical server. Using this well established domain we ablate training procedures (Section~\ref{sec:TrainingCaseStudy}) and deployment assumptions (Section~\ref{sec:DeployingCaseStudy}), to show how they can degrade performance.

    \textbf{Adversarial Malware Creation}
    We implement the malware detection and adversarial training environment AutoRobust~\cite{tsingenopoulos2024train} using publicly available data.\footnote{\url{https://www.kaggle.com/datasets/greimas/malware-and-goodware-dynamic-analysis-reports}}
    In this environment, the agent is presented with a detected malware sample, and must transform it so that it is misclassified. Rather than modifying the executable, the agent operates on a structured behavioral representation extracted from sandbox execution, consisting of system interactions such as file accesses, registry modifications, executed commands, and API usage. At each step, the agent edits or adds report entries to cross the decision boundary while maintaining a plausible behavioral trace. Using this environment, we disentangle the performance contributions of state/action space design from the learned policy, under both fully and partially observable settings. (Section~\ref{sec:EvaluatingCaseStudy}).

    \textbf{Web Security Testing} We consider two payload generation environments: Link~\cite{lee2022link} for Cross Site Scripting (XSS) and \sqirl~\cite{al2023sqirl} for SQL injection.
    The Link DRL agent uses a hand-crafted state-action space to iteratively build XSS payloads. We retrain Link on WAVSEP, a collection of deliberately vulnerable web-applications, following the settings from the original work. WAVSEP contains 32 XSS vulnerabilities and is used by Lee \etal~\cite{lee2022link}. We show the importance of environment modeling pitfalls by removing: the misaligned and partial observable states (Section~\ref{ModelingCaseStudy}). 
    The \sqirl\ DRL agent uses embeddings of SQL payloads, and statements to handle a dynamic action space, to generate payloads. We retrain \sqirl\ on its training benchmark, the \textsc{Smb}, which includes 20 training samples, and 10 testing samples (including 5 reproduced CVEs), following the original training and settings of the work. We show the effect of environment complexity by training \sqirl\ with and without experience bypassing sanitization; we then test them with both sanitized and unsanitized vulnerabilities (Section~\ref{sec:EvaluatingCaseStudy}).

\section{Pitfalls of Modeling Environments} \label{sec:pitfall-modeling}
    The foundation of applying \ac{DRL} is formulating a problem as an MDP~\cite{sutton2018reinforcement,chung1967markov}. As a result of problem diversity, cybersecurity domains typically lack standardized DRL environments necessitating bespoke \ac{MDP} formulations for each task. We identify three pitfalls in environment modeling (shown as \textcolor{pieGreenD}{green} in Figure~\ref{fig:drl-pitfalls-workflow}) that pose particular risks for security applications. First, an incomplete \textit{\msfull} (Section~\ref{sec:ms}) hinders reproducibility and obscures potential flaws in DRL-based approaches. Second, \textit{\mcfull} (Section~\ref{sec:mc}) issues arise when the underlying security task is incorrectly modeled as an MDP. Third, security problems often lack complete system information in practice; inadequately addressing this can lead to problems of \textit{\pofull} (Section~\ref{sec:po}). 

    \subsection{\msfull} \label{sec:ms}
 
        The \ac{MDP} defines the numerical optimization objective (\ie reward function) and structure of the environment (\ie state-action-transition dynamics) in which the agent learns from interaction~\cite{sutton2018reinforcement}. This is similar to the loss function and data distribution of input-output pairs in supervised learning.
        An unambiguous \ac{MDP} definition is thus essential for evaluating the suitability of any implementation or claims regarding agent performance~\cite{cavenaghi2023systematic}. However, we find that the \acp{MDP} specification in the cybersecurity literature often lacks comprehensive definitions. Common ambiguities in \ac{MDP} descriptions include high-level overviews and missing transition dynamics open to multiple interpretations.

        \begin{boxME}
            \begin{minipage}[c]{0.7\linewidth}
            \textbf{\msfull Pitfall} 
            
            The MDP is ill-defined, so that one or more components (state space, action space, reward function, and transition dynamics) remain unclear. 
            
            \end{minipage}
            \hspace{+0.1cm}
            \begin{minipage}[c]{0.25\linewidth}
            \centering
            \begin{tikzpicture}
                \pie[
                    radius=1,
                    hide number,
                    sum=100,
                    color={pieGreenD,pieGreenL,pieGrey}
                ]{
                    6/,
                    53/,
                    41/
                }
                \fill[draw=black,line width=1pt,fill=tcbcolback] (0,0) circle (0.7);
                \node at (0,0) {\textbf{M-MS}};
            \end{tikzpicture}
            \end{minipage}
        \end{boxME}

        \textbf{Prevalence of M-MS} Across the 66 papers reviewed, we observed fully unspecified \ac{MDP} definitions in 4 papers (\textcolor{textGreenD}{\textbf{6.1\%}}), partially unspecified definitions in 35 papers (\textcolor{textGreenL}{\textbf{53.0\%}}), and complete definitions in the remaining 27 papers (\textcolor{textGrey}{\textbf{40.9\%}}). Among the full and partial cases the distribution of underspecified components are as follows: the state space was the most frequent (22 papers, 33.3\%), followed by transitions (18 papers, 27.3\%), action space (14 papers, 21.2\%), and reward function (10 papers, 15.2\%). The high prevalence indicates a significant documentation issue in \drlsec literature, reflecting the inherent difficulty of translating complex security scenarios into fully specified environments~\cite{dulac-arnold_challenges_2019,bates2023reward}.
         
        \textbf{Security Implications of M-MS} The risks that ambiguous definitions can cause in real-world deployments is threefold.
        First, it obfuscates other pitfalls in the problem formulation, \eg state spaces that do not address partial observability allow attacks to go undetected despite causing harm to the system (see Section~\ref{sec:po}).
        Second, it hinders reproducibility, as attempts at reconstruction may yield different MDPs. For instance, `rewards for attack detection with penalties for false positives' can produce vastly different behaviors due to variations in relative reward magnitudes~\cite{basnet2025advanced}.
        Finally, assessing the real-world viability of the proposed solution becomes significantly harder. The lack of implementation details, \eg an underspecified state, can prevent assurance and auditing of potential attack vectors used in the MDP, thereby obfuscating the attack surface for defenders.

        \textbf{Recommendations for M-MS} Complete \ac{MDP} specification requires unambiguous definition of: (1) the state space, including available information, dimensionality, and semantic meaning of individual state values; (2) the action space, specifying all available actions and connection to the security task; (3) the transition function, describing how actions influence state changes, including stochasticity and terminal conditions; and (4) the reward function, providing the complete formulation with rationale for the contribution of each component to the overall objective, \eg~\cite{anderson2018learning,romdhana2022deep,al2023sqirl}. Crucially, assumptions, approximations, or simplifications must be explicitly discussed. When space constraints prevent full specification in the main text, this should be provided in appendices or supplementary materials, such as Lee \etal~\cite{lee2022link}. Clear \ac{MDP} definitions, combined with publicly available code, enable future work to avoid ambiguity, improve reproducibility, and expand on existing approaches.

    \subsection{\mcfull} \label{sec:mc}
        Cybersecurity applications typically lack pre-existing environments, requiring bespoke \ac{MDP} modeling, which can lead to subtle but fundamental errors in problem formulation. As a result, seemingly correct \ac{MDP}s may violate the Markov assumption or fail to encapsulate the real cybersecurity task. The Markov property (assumed to be true in MDPs) requires that future states and rewards depend only on the \emph{current} state and action~\cite{sutton2018reinforcement}. 
        However, this assumption breaks down in two cases: when environments rely on the \emph{entire} history, or when state transitions are \emph{independent} of the current state and action.
        Even when the Markov property is upheld, \mc can manifest through misalignment between the modeling of states, actions, or rewards, and the underlying security task. 
        Such misalignment is difficult to detect as policies can appear to perform well by achieving high rewards. For example, a network defense policy that locks all accounts due to misaligned rewards leads to a policy that has no real value.
        
        \begin{boxME}
            \begin{minipage}[c]{0.7\linewidth}
            \textbf{\mcfull Pitfall} 
            
            The modeled environment violates the Markov assumption or does not correctly encapsulate the actual \ac{MDP} underlying the security problem. 
            
            \end{minipage}
            \hspace{-0.01cm}
            \begin{minipage}[c]{0.25\linewidth}
            \begin{tikzpicture}
                \pie[
                    radius=1,
                    hide number,
                    color={pieGreenD,pieGreenL,pieGrey}
                ]{
                    25.8/,
                    9.1/,
                    65.2/
                }
                \fill[draw=black,line width=1pt,fill=tcbcolback] (0,0) circle (0.7);
                \node at (0,0) {\textbf{M-MC}};
            \end{tikzpicture}
            \end{minipage}
        \end{boxME}

        \textbf{Prevalence of M-MC} Across the 66 papers reviewed, we observed incorrect \ac{MDP} modeling in 17 papers (\textcolor{textGreenD}{\textbf{25.8\%}}), minor modeling errors in 6 papers (\textcolor{textGreenL}{\textbf{9.1\%}}), and valid modeling in the remaining 43 papers (\textcolor{textGrey}{\textbf{65.2\%}}). The combined 34.9\% of partially and fully present is primarily due to issues with state representation and transitions.
        While a general understanding of the \ac{MDP} framework is prevalent, unaddressed violations of the Markov property are still contained in a third of papers reviewed. This can be attributed to both the lack of standardized environments and the task being poorly suited to being formulated as an MDP.
        
        \textbf{Security Implications of M-MC} When the Markov property is violated or the task is misspecified as an MDP, agents can learn spurious correlations between unrelated decisions. This can result in policies that exploit artificial sequential patterns rather than genuine security relationships.
        For example, formulating a classification task as sequential decision-making induces nonexistent temporal dependencies between independent samples~\cite{McFadden2026DRMD}.
        Similarly, violations of the Markov property lead to biased and unstable learning, causing agents to exploit artifacts of the formulation rather than genuine structure in the problem. Separately, misaligned rewards can incentivize agents to maximize return through behaviors that undermine security objectives via \emph{reward hacking}~\cite{skalse_defining_2022,hessel_inductive_2019,abdelnabi2026measuring}. Together, these modeling errors produce brittle policies that appear effective in controlled experiments but fail unpredictably when deployed. Policies learned on an \ac{MDP} that does not reflect the real-world scenario lead to systems that are not only less effective but potentially more vulnerable.

        \textbf{Recommendations for M-MC} Modeling of security task as \ac{MDP}s must consider both theoretical and practical validity of the formulation. 
        First, the state transitions must reflect genuine sequential structure (next states are consequences of current state \emph{and} action) rather than artificially chaining independent decisions. For example, independent samples in a detection task should seek alternative methods such as deep contextual bandits~\cite{zhou2020neural,zhang2020neural} to avoid spurious correlations between states~\cite{McFadden2026DRMD}. 
        Secondly, \ac{MDP} components should be carefully defined such that: rewards provide learning signals to incentivize solving the security task rather than gaming the reward function; actions influence subsequent states to enable rational policy learning; and states include all available information of the environment without conflating details~\cite{foley2025apirl, bates2025less, terranova2024leveraging}. If complete state information is unavailable in practice, the problem should be formulated as a \ac{POMDP} (see Section~\ref{sec:po}) with appropriate constraints rather than forcing a potentially misaligned MDP representation.

    \subsection{\pofull} \label{sec:po}
        Security problems often inherently lack complete information, for example a defensive agent does not know the full range of attacks an adversary could employ. Such problems are described as \emph{partially observable}. In \acp{POMDP}, agents have only partial knowledge of an otherwise `hidden' state~\cite{hausknecht2015deep}. Treating a \ac{POMDP} as fully observable when it does not model full system dynamics, can lead to suboptimal behavior and systematic blind spots in the learned policy~\cite{simon_learning_2025}.
        Therefore, when an environment is partially observable, the implications must be explicitly analyzed and appropriately mitigated.

        \begin{boxME}
            \begin{minipage}[c]{0.7\linewidth}
            \textbf{\pofull Pitfall} 
            
            The environment is inherently partially observable (i.e. a \ac{POMDP}); however, it is treated as if it were fully observable \emph{without} mitigating the partial observability.
            
            \end{minipage}
            \hspace{-0.05cm}
            \begin{minipage}[c]{0.25\linewidth}
            \begin{tikzpicture}
                \pie[
                    radius=1,
                    hide number,
                    color={pieGreenD,pieGreenL,pieGrey}
                ]{
                    30.3/,
                    30.3/,
                    39.4/
                }
                \fill[draw=black,line width=1pt,fill=tcbcolback] (0,0) circle (0.7);
                \node at (0,0) {\textbf{M-PO}};
            \end{tikzpicture}
            \end{minipage}
        \end{boxME}

        \textbf{Prevalence of M-PO} Across the 66 papers reviewed, we identified clear partial observability issues in 20 papers (\textcolor{textGreenD}{\textbf{30.3\%}}) and lesser issues in 20 papers (\textcolor{textGreenL}{\textbf{30.3\%}}). Of these 40, 8 papers (12.1\%) explicitly acknowledge the presence and potential issues of partial observability, yet do not mitigate it. The remaining 26 papers (\textcolor{textGrey}{\textbf{39.4\%}}) either did not consider a partially observable task or directly accounted for it in their modeling. Such prevalence indicates that partial observability is a pervasive challenge in \ac{DRL} applications to cybersecurity. While the number of papers acknowledging the issue show that some awareness exists, it does not match the high prevalence of partial observability in cybersecurity tasks. Furthermore, this highlights the need to translate solutions from the broader \ac{DRL} community into cybersecurity contexts.
        
        \textbf{Security Implications of M-PO} 
        Unmitigated partial observability often manifests as a vulnerability \emph{after} training. 
        Consider an agent trained on incomplete information, if partial observability remains unmitigated, the agent may not infer the hidden dynamics of the environment required for effective performance.
        Therefore, agents can develop `blind spots' due to the unobservability of a system. Such policies may be optimal in simulation, but in the real-world they have a limited `key hole' view, leaving them vulnerable to malicious actors.
    
        \textbf{Recommendations for M-PO} 
        Partial observability may be inherent to security tasks which often feature incomplete, delayed, or deliberately concealed information. Addressing this in modeling is therefore essential for making robust decisions under uncertainty. In such cases it should be \emph{explicitly} formulated as a \ac{POMDP}, as done by Terranova \etal~\cite{terranova2024leveraging}. Agents can then use the (partial) information to infer the underlying, hidden state. Commonly this can be achieved by either encoding recent observations into the state representation, such as frame stacking used in Atari games~\cite{mnih2013playing}, or employing recurrent policies (\eg LSTM or Transformer)~\cite{hausknecht2015deep,meng2021memory} to maintain an internal memory. Other approaches aim to model the unobserved information, using techniques such as Hidden Markov Models~\cite{xu2007defending} or model-based approaches~\cite{kaiser2019model}, to mitigate the impact of \po. 

    \subsection{Case Study of Modeling Pitfalls} \label{ModelingCaseStudy}
    
        We use the Link XSS payload generation environment~\cite{lee2022link} to show how modeling decisions impact performance. Although the original environment contains valid Markovian transitions, its design introduces unnecessary partial observability and misalignment with the practical security task. 
        This arises from overlapping features in the state space, whereby multiple XSS payloads can be mapped to identical DRL states. For example, a single feature is used to indicate multiple HTML tags (\eg img, video, audio or svg tags).
        The unintentional pitfalls introduced by this design can be rectified through reformulation of the MDP.
        Thus, we consider two formulations of the MDP for this task: (1) \textit{Original}, as published in \cite{lee2022link} that have merged states;
        (2) \textit{Distinct States}, that unrolls the conflated payload features into distinct values, increasing the state dimensionality by 12.8\% from 47, to 53.
        Table~\ref{tab:modelingCaseStudy} shows the performance of these two MDPs over 20 runs.

        \input{Tables/Link-mini-table}

        \textbf{Modeling Correctness \& Partial Observability }
        Introducing \textit{Distinct States} for each possible payload aligns the MDP with the underlying security task avoiding \mc.
        Therefore, the agent can develop correct payloads by selecting actions and observing the \emph{distinct} impact they have on the current payload. Thus, removing the unnecessary partial observability, \po, from the environment. 
        This effect can be observed concretely in Table~\ref{tab:modelingCaseStudy} where including \emph{Distinct States} increases the percentage of vulnerabilities found over \emph{Original} by 10.1\% and the raising lower bound performance by 17.3\%. 
        Therefore, through minor changes in the state space to remove the \mc and \po pitfalls, we can observe significant improvements in performance.
        
        \textbf{MDP Specification} Although originally defined clearly, the performance difference between MDP formulations in Table~\ref{tab:modelingCaseStudy} demonstrates that when components are left unclear, \msfull can severely impact final performance.

    \section{Pitfalls of Training Agents} \label{sec:pitfall-training}
    
    Unlike supervised learning, where training typically converges given sufficient data, \ac{DRL} training is inherently more variable and sensitive to numerous factors including: hyperparameter choices, random initialization, sparsity of reward, and the stochastic nature of exploration~\cite{adkins2024method, henderson2018deep,patterson2024empirical}. We identify three pitfalls in training that can compromise the trustworthiness of DRL-based systems in security. First, \textit{\hrfull} (Section~\ref{sec:hr}) issues arise when critical implementation details necessary for reproduction and validation are omitted. Second, when \textit{\vafull} (Section~\ref{sec:va}) is not performed across multiple training runs, the reliable performance of the proposed approach is in question. Third, when \textit{\pcfull} (Section~\ref{sec:pc}) is not achieved, due to premature training termination, it often leads to systems with inconsistent and unpredictable behavior when deployed. 

    \subsection{\hrfull} \label{sec:hr}
        Hyperparameter and architectural choices for \ac{DRL} agents can have significant impacts on agent training, and final performance, due to the high sensitivity of \ac{DRL} to these design choices~\cite{patterson2024empirical, eimerhyperparameters}.
        We find that \drlsec research frequently fails to report \ac{DRL} hyperparameters, including: agent architectures, random seeds, discount factors, and specific DRL algorithm settings (\eg clipping coefficient for PPO~\cite{PPO}). This information is essential to enable reproducibility and fair comparison between proposed approaches. 

        \begin{boxTA}
        \begin{minipage}[c]{0.7\linewidth}
        \textbf{\hrfull Pitfall} 
        
        The hyperparameters governing training and agent architecture of an approach are not reported.
        
        \end{minipage}
        \hspace{-0.09cm}
        \begin{minipage}[c]{0.25\linewidth}
        \begin{tikzpicture}
            \pie[
                radius=1,
                hide number,
                color={pieBlueD,pieBlueL,pieGrey}
            ]{
                31.8/,
                36.4/,
                31.8/
            }

            \fill[draw=black,line width=1pt,fill=tcbcolback] (0,0) circle (0.7);
        
            \node at (0,0) {\textbf{T-HR}};
        \end{tikzpicture}
        \end{minipage}
        \end{boxTA}

        \textbf{Prevalence of T-HR} Across the 66 papers reviewed, we observed the complete absence of hyperparameters in 21 papers (\textcolor{textBlueD}{\textbf{31.8\%}}), partial reporting of hyperparameters in 24 papers (\textcolor{textBlueL}{\textbf{36.4\%}}), and complete hyperparameter reporting in the remaining 21 papers (\textcolor{textGrey}{\textbf{31.8\%}}). The high rate of missing hyperparameters, is significant given their importance for \ac{DRL}. This omission fundamentally undermines reproducibility and hinders the advancement of \drlsec applications. 
        
        \textbf{Security Implications of T-HR} Since minor variations in design choices can have a disproportionate effect on agent behavior, under-specifying hyperparameters makes results challenging to verify~\cite{cavenaghi2023systematic,patterson2024empirical}. Hyperparameter transparency is not only a matter of scientific rigor, but a prerequisite for establishing reliable and robust DRL cybersecurity applications that can perform tasks effectively. 
    
        \textbf{Recommendations for T-HR}
        Complete hyperparameter reporting is necessary for reproducible, trustworthy research in cybersecurity as DRL algorithms are highly sensitive to hyperparameter choices~\cite{adkins2024method, henderson2018deep}. At minimum, publications should report: (1) \textit{algorithm-specific hyperparameters}, \eg learning rate, discount factor $\gamma$,  batch size, replay buffer size, and exploration parameters such as $\epsilon$; (2) \textit{neural network architecture details}, \eg layer types, dimensions, activation functions, normalization techniques; (3) \textit{optimization settings}, \eg optimizer type, learning rate, gradient clipping coefficients; and (4) \textit{training procedures}, \eg total timesteps, update frequency, number of environments for parallel training. When space constraints prevent inclusion in the main text, it should be provided in appendices or supplementary materials, \eg \cite{tsingenopoulos2022captcha,terranova2024leveraging,caminero2019adversarial}.

    \subsection{\vafull} \label{sec:va}
        Variance in training is an intrinsic property of \ac{DRL} that can arise from a number of factors including: stochastic environments, exploration noise, and probabilistic transitions~\cite{patterson2024empirical}. Therefore, training a single policy risks over-interpreting randomness or potential outlier runs as the policy is not representative of aggregate performance. Training multiple policies allows the use of statistical metrics to evaluate the performance and variance of an approach.

        \begin{boxTA}
            \begin{minipage}[c]{0.7\linewidth}
            \textbf{\vafull Pitfall} 
            
            The inherent variance of \ac{DRL} training is not investigated through explicit analysis of performance over multiple policies.
            
            \end{minipage}
            \hfill
            \begin{minipage}[c]{0.25\linewidth}
            \begin{tikzpicture}
                \pie[
                    radius=1,
                    hide number,
                    color={pieBlueD,pieBlueL,pieGrey}
                ]{
                    57.6/,
                    9.1/,
                    33.3/
                }
                \fill[draw=black,line width=1pt,fill=tcbcolback] (0,0) circle (0.7);
                \node at (0,0) {\textbf{T-VA}};
            \end{tikzpicture}
            \end{minipage}
        \end{boxTA}
 
        \textbf{Prevalence of T-VA} Across the 66 papers reviewed, we observed no discussion of multiple training `runs' or variance analysis in 38 papers (\textcolor{textBlueD}{\textbf{57.6\%}}), partial consideration of variance in 6 papers (\textcolor{textBlueL}{\textbf{9.1\%}}), and multiple runs with variance analysis in the remaining 22 papers (\textcolor{textGrey}{\textbf{33.3\%}}).  Indeed, two thirds of reviewed studies (44 papers, 66.7\%) insufficiently consider variance.
        This lack of analysis obscures genuine methodological advances by reporting results that may not be typical of performance, but instead the result of favorable randomness (\eg seed values) in training~\cite{patterson2024empirical}. 
        
        \textbf{Security Implications of T-VA}
        Favorable initial conditions, or `beginner's luck', a single run can over- or underestimate the perceived performance beyond the realistic mean of multiple runs. 
        This is particularly unsafe in security contexts where consistent and reliable performance is crucial. 
        Dismissing an approach based on a single bad run that can perform well \emph{on average}, risks rejecting a valuable security solution. 
        Alternately, agents that over-promise and under-deliver on average, expose end users to unnecessary risk.
    
        \textbf{Recommendations for T-VA}
        As a lone trained agent is insufficient to establish representative performance of that agent~\cite{Agarwal}, it is essential to train and evaluate multiple trained policies and report the statistical variance. Yet, there is no canonical number of policy runs $N$ to use for evaluation.
        While it has been argued that $N\geq20$ is a reasonable mean~\cite{agarwal2022reincarnating, Colas}, complex environments may require $N\geq50$ for robust 95\% confidence intervals~\cite{Agarwal}, and often \ac{DRL} papers report $N\leq5$~\cite{PPO, Dabney, osband, Chen}. Consequently, we recommend a minimum of $N\geq5$, and encourage higher runs ($N\geq20$ or $N\geq50$) for stronger statistical claims~\cite{patterson2024empirical}. 
        To better represent the variance in agent behavior, statistical metrics must also be reported~\cite{Agarwal, chan2019measuring, patterson2024empirical}, for example: 95\% confidence intervals, inter-quartile range (IQR), and risk-sensitive measures such as Conditional Value at Risk (CVaR). 
        Variance should be reported within and between training episodes, and certainly for the final policies during testing~\cite{chan2019measuring}. Examples of such testing can be seen in~\cite{de2024guideenricher,erdHodi2021simulating, bates2023reward}. 
        In cybersecurity, reliability directly impacts security posture. Transparent variance reporting enables informed risk assessments from practitioners, preventing the reliance on potentially misleading results that will not reflect performance in deployment.

    \subsection{\pcfull} \label{sec:pc}
        Training in \ac{DRL} should result in a policy that has converged to a \emph{stable} strategy, even if sub-optimal. Determining convergence is a difficult and open problem in DRL~\cite{PPO,osband,Chen}.
        However, in \drlsec literature the training length, a key aspect in investigating convergence, is either frequently not reported or unjustified.
        Without presenting reward or performance metrics during training, it is impossible to determine policy convergence or premature training termination.

        \begin{boxTA}
            \begin{minipage}[c]{0.7\linewidth}
            \textbf{\pcfull Pitfall} 
            
            The convergence of the policy is not demonstrated prior to the termination of training.
            
            \end{minipage}
            \hspace{-0.01cm}
            \begin{minipage}[c]{0.25\linewidth}
            \begin{tikzpicture}
                \pie[
                    radius=1,
                    hide number,
                    color={pieBlueD,pieBlueL,pieGrey}
                ]{
                    42.4/,
                    28.8/,
                    28.8/
                }
                \fill[draw=black,line width=1pt,fill=tcbcolback] (0,0) circle (0.7);
                \node at (0,0) {\textbf{T-PC}};
            \end{tikzpicture}
            \end{minipage}
        \end{boxTA}
 
        \textbf{Prevalence of T-PC} Across the 66 papers reviewed, we observed no demonstration of convergence in 28 papers (\textcolor{textBlueD}{\textbf{42.4\%}}), partial demonstration or discussion in 19 papers (\textcolor{textBlueL}{\textbf{28.8\%}}), and demonstration of convergence in the remaining 19 papers (\textcolor{textGrey}{\textbf{28.8\%}}). While many papers discuss the reasoning for the length of training used or mention convergence, explicit demonstration is often missing. The gap in convergence reporting suggests that although the need for convergence may be understood, the concrete demonstration of policy convergence and stability is often under-emphasized. In this way, the reliability of the learned behaviors is uncertain.
        
        \textbf{Security Implications of T-PC} Without demonstrating convergence, it is difficult to establish if an agent has learned a truly robust policy or if premature termination has left gaps in its decision-making for certain scenarios~\cite{wuAdversarial2021}. 
        Non-converged policies pose an increased risk as they behave unpredictably, which can be exacerbated in rare edge cases~\cite{sun2020stealthy}. 
        In such scenarios, even otherwise optimal agents can behave erratically, leading to clear vulnerabilities in deployment~\cite{gleave2019adversarial,sun2020stealthy}.
    
        \textbf{Recommendations for T-PC}
        The convergence of policies should be evaluated over the learning process to reveal training dynamics, such as instability or temporary performance plateaus before final convergence. 
        We recommend the inclusion of figures showing episodic reward or task-specific metrics as clear evidence of stabilization before training termination~\cite{patterson2024empirical, machado2018revisiting}.  
        To ensure convergence is consistent, measures of training variance should be included as discussed in Section~\ref{sec:va}. 
        This approach aids in the assessment of policies that represent genuinely converged behaviors, and not merely snapshots of incomplete learning that may fail unpredictably, as can be seen in~\cite{li2022alphaprog,luo2023autocat,wang2020coordinated}. 

    \subsection{Case Study of Training Pitfalls} \label{sec:TrainingCaseStudy}

        To evaluate the impact of pitfalls in policy training, we use MiniCAGE~\cite{emerson_cyborg_2024}, described in Section~\ref{sec:case-study-envs}. Specifically, we evaluate four hyperparameter settings for \hr, investigate variance and convergence for \va and \pc, respectively.

        \begin{figure}[t]
            \centering
            \includegraphics[width=\linewidth]{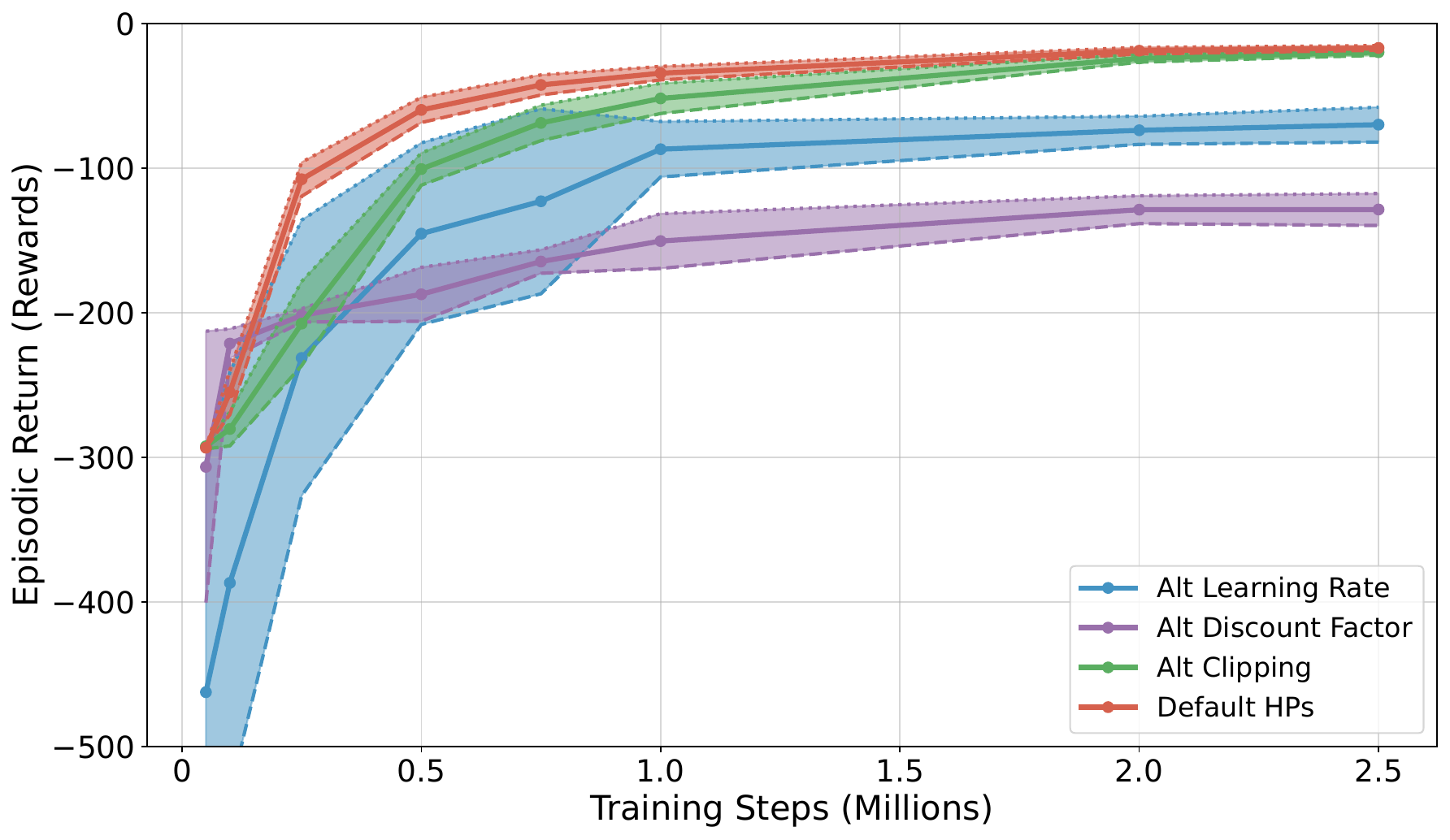}
            \caption{Training performance of different hyperparameter sets, across 20 runs using 95\% CI for variance. Zero is the theoretical max score in MiniCAGE.}
            \label{fig:CAGE-Training}
        \end{figure}

        \textbf{Impact of Hyperparameters} In Figure~\ref{fig:CAGE-Training}, we show the performance of a default hyperparameter set with three modifications: `Default HPs' (\textcolor{textRedD}{red}) represents the default hyperparameters in Stable Baselines 3\footnote{https://github.com/DLR-RM/stable-baselines3} for PPO; `Alt Learning Rate' (\textcolor{textBlueD}{blue}) modifies the learning rate; `Alt Discount Factor' (\textcolor{textPurpleD}{purple}) modifies the discount factor for future rewards; and `Alt Clipping' (\textcolor{textGreenD}{green}) modifies PPO's loss clipping. We examine the final performance at 2.5 million steps and see how changing a single hyperparameter results in drastically different performance, highlighting how hyperparameter transparency is needed to ensure reproducibility.

        \textbf{Analyzing Variance} For each evaluation in Figure~\ref{fig:CAGE-Training}, we plot both the mean performance of twenty agents and the 95\% Confidence Interval (CI) upper and lower bounds. We can see the potential impact of not performing variance analysis. Concretely, at 500k the upper CI bound of the `Alt Learning Rate' (\textcolor{textBlueD}{blue}) appears to be the second best performing agent. However, we see from the variance that an agent at this upper CI bound is not representative of the (lower) mean performance.  By evaluating over multiple runs and reporting 95\% CI, we obtain a more accurate picture of actual performance.

        \textbf{Demonstrating Policy Convergence} Figure~\ref{fig:CAGE-Training} also plots the policy performance (episodic rewards) over training. The `Default HPs' (\textcolor{textRedD}{red}) and `Alt Clipping' (\textcolor{textGreenD}{green}) settings show relative convergence at 2.0-2.5 million steps where plateauing performance curves and minor variance can be seen. In contrast, the `Alt Learning Rate' (\textcolor{textBlueD}{blue}) and `Alt Discount Factor' (\textcolor{textPurpleD}{purple}) are still increasing in performance with higher variance at 2.5 million steps. Plots such as this, in conjunction with variance analysis, can be used to effectively demonstrate convergence and motivate training length chosen.

\section{Pitfalls of Evaluating Agents} \label{sec:pitfall-evaluation}
    \ac{DRL} evaluations present unique challenges compared to traditional machine learning. We identify three pitfalls in agent evaluation that can undermine DRL-based approaches in cybersecurity. First, \textit{\amfull} (Section~\ref{sec:am}) issues occur when \ac{DRL} is used without adequate justification over simpler domain-specific methods. Second, \textit{\gafull} (Section~\ref{sec:ga}) issues arise when performance gains are due to the basic agent capabilities rather than the learned policy itself. Third, \textit{\ecfull} (Section~\ref{sec:ec}) creates misleading results when training and evaluation occur in overly simplified settings that do not capture real-world complexity.

    \subsection{\amfull} \label{sec:am}
        Motivating an applied approach and comparing to established baselines are standard scientific practice. Therefore, the application of DRL must be motivated over existing approaches both empirically (through experimental comparison) and theoretically (by justifying the value of DRL-based approaches).
        Yet, direct and fair comparisons to existing approaches in the domain can be infeasible when reformulating through a \ac{DRL} paradigm. For example, Goel~\etal~\cite{goel2024optimizing} design \emph{dynamic} active directory defense as a two player attacker-defender game, as opposed to static detection, thereby making comparison with existing approaches unreasonable.  
        It is therefore crucial in such cases to theoretically motivate the necessity of \ac{DRL} over existing formulations (for example see Section~\ref{sec:EvaluatingCaseStudy}).

        \begin{boxEA}
            \begin{minipage}[c]{0.7\linewidth}
                \textbf{\amfull Pitfall} 
                
                The application of \ac{DRL} over traditional methods is not justified empirically or theoretically.
    
                \end{minipage}
                \hspace{+0cm}
                \begin{minipage}[c]{0.25\linewidth}
                \begin{tikzpicture}
                    \pie[
                        radius=1,
                        hide number,
                        color={piePurpleD,piePurpleL,pieGrey}
                    ]{
                        12/,
                        32/,
                        56/
                    }
                
                    \fill[draw=black,line width=1pt,fill=tcbcolback] (0,0) circle (0.7);
                
                    \node at (0,0) {\textbf{E-AM}};
                \end{tikzpicture}
                \end{minipage}
            \end{boxEA}

        \textbf{Prevalence of E-AM} Across the 66 papers reviewed, we found that 8 papers (\textcolor{textPurpleD}{\textbf{12.1\%}}) provide neither empirical or theoretical motivation for their use of \ac{DRL}; 21 papers (\textcolor{textPurpleL}{\textbf{31.8\%}}) offered only one of the forms of motivation; and 37 papers (\textcolor{textGrey}{\textbf{56.1\%}}) motivated in both forms. The majority of partial justification (17 papers, 25.8\%) arises from only having theoretical motivation. The absence of appropriate baselines may stem from the difficulty of aligning reformulated tasks to prior work, or a focus shift from investigating concrete improvements to the novelty of applying \ac{DRL}.
        
        \textbf{Security Implications of E-AM} The application of \ac{DRL} in security is relatively under-explored in comparison to other learning paradigms. Replacing extensively studied methods with DRL, besides the other pitfalls we explore here, runs the risk of introducing novel attack surfaces~\cite{gleave2019adversarial,sun2020stealthy,shereen2025one}. Therefore, the benefits of a DRL-based approach should surpass engineering in-domain approaches to replicate the sequential nature of \ac{DRL} environments.
     
        \textbf{Recommendations for E-AM} 
        While \ac{DRL} can perform a wide variety of complex tasks, much like other ML paradigms it is not suitable for all tasks. As a result, applying DRL to a novel task should be justified theoretically and, when possible, empirically. 
        When baselines are available for comparison, they should be used to demonstrate the advantage of \ac{DRL}, in addition to the motivation for applying DRL in the first place~\cite{tong2020finding,chen2024llm,yang2020patchattack}. 
        When baselines are not available, motivation \emph{must} demonstrate the \textit{necessity} of re-formulating to a \ac{DRL} task~\cite{terranova2024leveraging,kvasov2023simulating}.

    \subsection{\gafull} \label{sec:ga} 
        As discussed in Section~\ref{sec:ms} framing a security task as an MDP is the foundation of applying \ac{DRL}. However, engineering a functional environment entails additional design choices beyond state representation, particularly in how actions are translated into concrete effects in the cybersecurity task~\cite{patterson2024empirical}. These artifacts can introduce performance gains that overstate the actual contribution of the learned agent.
    
    \begin{boxEA}
    \begin{minipage}[c]{0.7\linewidth}
                \textbf{\gafull Pitfall} 

                Improvements stem from additional information or capabilities provided in MDP design rather than the learned policy itself.
    
    \end{minipage}
    \hspace{+0.12cm}
    \begin{minipage}[c]{0.25\linewidth}
    \begin{tikzpicture}
        \pie[
            radius=1,
            hide number,
            color={piePurpleD,piePurpleL,pieGrey}
        ]{
            23/,
            27/,
            50/
        }
    
        \fill[draw=black,line width=1pt,fill=tcbcolback] (0,0) circle (0.7);
    
        \node at (0,0) {\textbf{E-GA}};
    \end{tikzpicture}
    \end{minipage}
    \end{boxEA}

        \textbf{Prevalence of E-GA} Across the 66 papers reviewed, we found that 15 papers (\textcolor{textPurpleD}{\textbf{22.7\%}}) did not clearly demonstrate that the agent improved over baseline environmental performance; 18 papers (\textcolor{textPurpleL}{\textbf{27.3\%}}) demonstrated partial evaluation of where performance gains come from; and 33 papers (\textcolor{textGrey}{\textbf{50.0\%}}) adequately showed the contribution of the learned policy. Notably, 6 of the 15 papers fully exhibiting this pitfall \emph{explicitly} acknowledge it but do not attempt to disentangle the sources of improvement. 
        
        \textbf{Security Implications of E-GA}  
        Misrepresentative gains can create false confidence in a proposed \ac{DRL} approach. Even when an approach outperforms state-of-the-art, gains may stem primarily from baseline environmental capabilities rather than the learned policy, undermining conclusions about the necessity and benefit of DRL~\cite{henderson2018deep}. Such agents can give a false sense of robustness as their learned policy is not the primary source of effectiveness, making it easier for an adversary to attack~\cite{wuAdversarial2021,gleave2019adversarial}. Thus such improvements reflect stronger baseline design rather than effective learning, and could be replicated by other agents given the same capabilities.

        \textbf{Recommendations for E-GA}
        The value of the learned policy can be disentangled from environmental artifacts through two kinds of ablation. 
        First, evaluating performance under random action sampling isolates the contribution of policy learning and establishes a baseline performance in the environment~\cite{patterson2024empirical}. As a result, improvements that arise solely from state-action engineering should be reported separately from the trained policy performance, for example: ~\cite{li2023innes,bottinger2018deep}.  
        Second, comparisons between different approaches (both \ac{DRL} and other-paradigms) should be performed with equivalent information and capabilities (\eg state–action spaces), as shown in ~\cite{deng2024ransomware,wang2020coordinated}. 
        This isolates the improvements introduced by MDP engineering, such as expanded information in the states and enhanced capabilities of the actions.

    \subsection{\ecfull} \label{sec:ec}
        Environments provide an interactive interface for a \ac{DRL} agent, translating agent actions to concrete security outcomes, and providing meaningful feedback, for example mutating payloads and observing their effects~\cite{lee2022link,al2023sqirl,foley2022haxss,mcfadden2024wendigo}. 
        In practice, implementing real-world environments is often complex and highly variable. The same security problem can be approached in a number of ways~\cite{collyer_acd-g_2022,msft:cyberbattlesim,bates2023reward,emerson_cyborg_2024,applebaum_bridging_2022}, using environmental abstractions, simulations, or simplifications~\cite{msft:cyberbattlesim,wang2020coordinated,erdHodi2021simulating,landen2022dragon}. 
        Many cybersecurity environments do not demonstrate a strong real-world correspondence and, consequently, agent performance has limited relevance to the fundamental security task~\cite{moore_fundamental_2025,dulac-arnold_challenges_2019}.

\begin{boxEA}
\begin{minipage}[c]{0.7\linewidth}
            \textbf{\ecfull Pitfall} 
            
            Training and evaluation occur in an environment that is contrived or overly simplified.

\end{minipage}
\hspace{+0.03cm}
\begin{minipage}[c]{0.25\linewidth}
\begin{tikzpicture}
    \pie[
        radius=1,
        hide number,
        color={piePurpleD,piePurpleL,pieGrey}
    ]{
        18/,
        23/,
        59/
    }

    \fill[draw=black,line width=1pt,fill=tcbcolback] (0,0) circle (0.7);

    \node at (0,0) {\textbf{E-EC}};
\end{tikzpicture}
\end{minipage}
\end{boxEA}

        \textbf{Prevalence of E-EC} Across the 66 papers reviewed, we found that 12 papers (\textcolor{textPurpleD}{\textbf{18.2\%}}) evaluated their approaches in overly simplified or contrived environments; 15 papers (\textcolor{textPurpleL}{\textbf{22.7\%}}) made some effort to improve environmental realism; and 39 papers (\textcolor{textGrey}{\textbf{59.1\%}}) evaluated on sufficiently realistic environments. 
        The 40.7\% of papers with this pitfall highlight that implementing real-world environments is non-trivial. 
         
        \textbf{Security Implications of E-EC}  Agents trained in overly simplified environments may learn effective policies, but these may be incapable of transferring to, or learning in, real environments~\cite{skalse_defining_2022,hessel_inductive_2019}. Indeed, the transfer from simulated to real environments (commonly referred to as the `sim-to-real' gap~\cite{nyberg2023training}) can lead to catastrophic failures. A concrete example of this is \emph{reward hacking} where agents learn to exploit flaws in environmental modeling to maximize rewards, leading to policies that appear optimal but provide no genuine security value~\cite{skalse_defining_2022,nikishin_primacy_2022}. These flaws become particularly dangerous in security contexts where the stakes of failure are high: organizations may deploy systems based on promising laboratory results only to discover that they underperform or are vulnerable in the real-world~\cite{9308468}.
        
        \textbf{Recommendations for E-EC}       
        As a result of transferability issues from the sim-to-real gap, it is best practice to train and evaluate DRL in the most realistic environment feasible. 
        Applying \ac{DRL} to security problems should follow the evaluation best practices from that specific security domain, \eg,~\cite{anderson2018learning,bottinger2018deep}.
        For instance, in security domains where real-world data or systems are used (\eg web security testing often using live applications), \ac{DRL} should not preclude such an evaluation. 
        However, in some domains the use of real-world data or systems is challenging or infeasible.
        In such cases benchmarks should be used to demonstrate impact. 
        For example, \ac{ACD} research relies on a handful of benchmark simulators~\cite{cage_challenge_2,msft:cyberbattlesim,yawningtitan}, while fuzzing research recommends benchmarks containing known vulnerabilities~\cite{klees_evaluating_2018,schloegel_sok_2024}.
        
    \subsection{Case Study of Evaluation Pitfalls} \label{sec:EvaluatingCaseStudy}
        Here we consider the three pitfalls we identified in evaluating agents: first, the attribution of gains in the AutoRobust environment; second, the theoretical motivation of \ac{DRL} in adversarial malware generation; third, environment complexity comparisons in the \sqirl\ environment for SQL injection.

        \begin{figure}[t]
        \centering
        \includegraphics[width=1\linewidth]{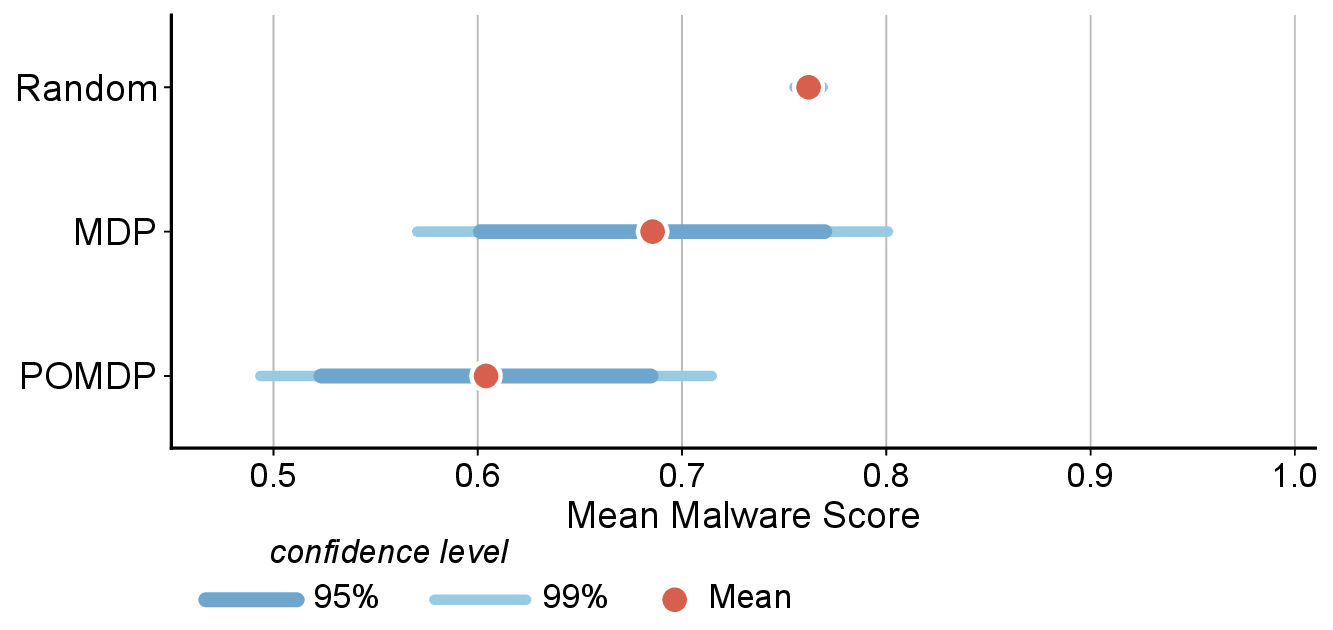}
        \caption{Comparison of mean with 95/99\% CI performance for trained and random policies in AutoRobust across 20 runs with 20 evaluation episodes each.}
        \label{fig:autorobust}
        \end{figure}
        
        \textbf{Identifying the Source of Gains}. To assess the impact of the \ga pitfall, the goal in the AutoRobust environment (modeled after~\cite{tsingenopoulos2024train}) is to disentangle the contributions of different elements in the approach; see the results in Figure~\ref{fig:autorobust}.
        First, we train and evaluate policies with full access to the internal state of the malware detection model, referred to as \emph{MDP}. 
        Second, we consider policies without internal state access, named \emph{POMDP}, to evaluate the performance of learning under a more restrictive and realistic threat model.
        Finally, we evaluate \emph{Random} policies that sample actions at random (as recommended in Section~\ref{sec:ga}), to isolate the intrinsic efficacy of the action space itself.
        There are two key takeaways from results in Figure~\ref{fig:autorobust}. First, full observability does not guarantee better performance, as agents can struggle to learn the task even with fully observable information. Second, while trained policies (MDP \& POMDP) have the \textit{potential} to achieve better performance, the random baseline is already strong. This demonstrates the importance of \ga as the action space alone provides a large portion of overall performance.

        \textbf{Motivating the Application}.  
        As an example of theoretical motivation, in the domain of adversarial malware creation, \ac{DRL} has several key capabilities over existing gradient-based attacks.
        The increasing complexity of AI-based systems makes gradient-based methods intractable. Concretely, gradient-based methods assume a fixed input size, limiting exploration to only a very small subset of the actual transformations an adversary can make~\cite{zou2023universal}.
        Similarly in malware, gradient-based perturbations on the feature representation do not have a consistent and reliable method for being mapped to valid changes in the problem-space, \ie the actual program. 
        In contrast, DRL can operate \textit{directly} on malware samples, bypassing the need for feature to problem-space mapping.
        Furthermore, given an adversary with a well defined sets of actions, at convergence DRL can additionally confer probabilistic guarantees on the level of vulnerability/robustness~\cite{tsingenopoulos2024train}.
        
       \input{Tables/SQiRL_EC-CS-Table} 

       \textbf{Changing Environment Complexity} To show the effect of \ec we use the \textsc{Smb} benchmark from \sqirl~\cite{al2023sqirl}. 
       Using the \textsc{Smb} we train two policies, one on the default \textsc{Smb} with a \emph{Combined} sanitized and non-sanitized vulnerability set, and one policy with the same vulnerabilities without sanitization (\emph{Non-Sanitized}).
       We evaluate both trained policies on the \textsc{Smb} test set of sanitized and non-sanitized vulnerabilities; Table~\ref{tab:sqirl} shows the percentage of vulnerabilities found across 20 runs with 95\% CI.
       We see both the \textit{Combined} and \textit{Non-Sanitized} agents are capable of discovering vulnerabilities in the simplified non-sanitized settings.
       However, when changing the evaluation set to the more challenging \emph{Sanitized} vulnerabilities we see a 47.4\% drop in vulnerabilities found for the \emph{Combined} agent, with a further 11.3\% for the \textit{non-sanitized} agent.
       This demonstrates how simplified evaluation environments can inflate performance, producing promising results that fail to translate into real-world capabilities.

\section{Pitfalls of Deploying Agents} \label{sec:pitfall-deployment}
    
    Avoiding pitfalls in modeling, training, and evaluation does not guarantee an agent's reliable performance in the complexities and uncertainties of real-world systems. Deployment introduces additional challenges that test the practical feasibility, robustness, and adaptability of DRL-based approaches. We identify two pitfalls in agent deployment that hinder real-world performance. First, \textit{\uafull} (Section~\ref{sec:ua}) issues arise when agent design or environment assumptions cannot be satisfied in practice. Second, \textit{\nsfull} (Section~\ref{sec:ns}) problems occur when the evolving nature of security environments are not adequately captured.

    \subsection{\uafull} \label{sec:ua}
        Modeling and implementing a security task with \ac{DRL} can introduce unrealistic assumptions that violate practical deployment constraints. While similar to \mcfull, an environment suffering from \ua can be theoretically valid, but invalid in practice. For example, assuming perfect knowledge of the network state in \ac{ACD}, unlimited query budgets for black-box attacks, unrealistic attack capabilities~\cite{pierazzi2020intriguing}, or continued access to ground-truth for rewards in evaluation. While these violations make a problem tractable for research, they can make the proposed approach infeasible, impractical, or unusable in realistic security deployments.

        \begin{boxDA}
        \begin{minipage}[c]{0.7\linewidth}
            \textbf{\uafull Pitfall} 
                    
            The environment or agent training requirements cannot be satisfied in real-world applications.
        
        \end{minipage}
        \hspace{+0.05cm}
        \begin{minipage}[c]{0.25\linewidth}
        \begin{tikzpicture}
            \pie[
                radius=1,
                hide number,
                color={pieRedD,pieRedL,pieGrey}
            ]{
                15/,
                14/,
                71/
            }
        
            \fill[draw=black,line width=1pt,fill=tcbcolback] (0,0) circle (0.7);
        
            \node at (0,0) {\textbf{D-UA}};
        \end{tikzpicture}
        \end{minipage}
        \end{boxDA}
        
        \textbf{Prevalence of D-UA} Across the 66 papers reviewed, we found that 10 papers (\textcolor{textRedD}{\textbf{15.2\%}}) contained unrealistic assumptions that would prevent real-world deployment (\eg network defense on abstract graph structures); 9 papers (\textcolor{textRedL}{\textbf{13.6\%}}) suffered from minor or partially unrealistic assumptions (\eg requiring real-time high quality labeled data in live deployments); and 47 papers (\textcolor{textGrey}{\textbf{71.2\%}}) operated under realistic deployment assumptions. Across the 19 papers beset by this pitfall, 6 papers (9.1\%) acknowledged the unrealistic nature of their assumptions and an awareness of practical deployment challenges. 
        We observe that assumptions regarding information access and capabilities are commonly relaxed when translating security problems to \ac{DRL}. Importantly, if constraint relaxation is unavoidable, a clear discussion should be provided to understand the limitations and trade-offs introduced. 
        
        \textbf{Security Implications of D-UA} \uafull can create overconfidence in the capabilities of a \ac{DRL}-based approach, which may not be directly evident during deployment~\cite{moore_fundamental_2025}.
        Such a system could be ineffective, under-performing, or even exploited, due to the \textit{disconnect} between research and actual deployment~\cite{gleave2019adversarial,nikishin_primacy_2022}.
        Furthermore, in cases of hard constraints like privileged data access or strict query budgets, the whole approach might be rendered invalid.
        
        \textbf{Recommendations for D-UA} To ensure practical relevance and transparency we recommend designing evaluations to mirror real-world problems.
        However, training effective \ac{DRL} agents often requires training over \emph{millions} of timesteps and data points~\cite{wurman_outracing_2022,openai_dota_2019}. 
        In some cases, sample-efficient or offline methods could minimize the number of data points required.
        Equally, other cases may require \emph{reasonable} assumptions regarding agent knowledge, capabilities, and operational conditions, including the information and resources that are available ~\cite{tong2020finding,wang2021crafting}.
        Realistic deployment settings should be used by constraining observation/action spaces, and enforcing practical limits on query budgets and computation.
        In cases where such assumptions are unavoidable, ablations should show performance and robustness under constrained and noisy conditions~\cite{le2024exploration,zhang2023moving}.
        Robustness and generality of results can be further strengthened through adversarial stress-testing to simulate operational uncertainty and attacker adaptation~\cite{tsingenopoulos2024train}.
        Finally, every study should include a brief summary of deployment requirements, such as privileges and training budgets.

    \subsection{\nsfull} \label{sec:ns}
        \ac{DRL} assumes static environment dynamics over time, violation of this assumption can lead to poor agent performance and generalization in deployment. Unfortunately, many security tasks are inherently non-stationary~\cite{Tesseract,kan2024tesseract}, including: evolving defenders~\cite{rashid2023malprotect,mcfadden2023poster,mcfadden2024impact,chen2020stateful}, adaptive adversaries~\cite{feng2023stateful, tsingenopoulos2023adversarial}, changing network topologies, updated software systems, and shifting attack methodologies. Thus, it is critical that such non-stationary components are integrated into the dynamics of the environment during training.

            \begin{boxDA}
            \begin{minipage}[l]{0.7\linewidth}
                \textbf{\nsfull Pitfall} 
                The environment does not effectively mitigate the inherent non-stationarity of the underlying cybersecurity task.
            \end{minipage}
            \hspace{0cm}
            \begin{minipage}[r]{0.25\linewidth}
            \begin{tikzpicture}
                \pie[
                    radius=1,
                    hide number,
                    color={pieRedD,pieRedL,pieGrey}
                ]{
                    35/,
                    20/,
                    45/
                }
                \fill[draw=black,line width=1pt,fill=tcbcolback] (0,0) circle (0.7);
                \node at (0,0) {\textbf{D-NS}};
            \end{tikzpicture}
            \end{minipage}
            \end{boxDA}
            
        \textbf{Prevalence of D-NS} Across the 66 papers reviewed, we found that 23 papers (\textcolor{textRedD}{\textbf{34.9\%}}) failed to mitigate the inherent non-stationarity of their cybersecurity tasks; 13 papers (\textcolor{textRedL}{\textbf{19.7\%}}) considered some of the non-stationarity present in the task; and 30 papers (\textcolor{textGrey}{\textbf{45.5\%}}) did not have, or adequately handled, the non-stationary aspects of their chosen task. The most common unaddressed non-stationarity across the papers was the behavior of other actors in the environment, \eg assumed fixed attacker in \ac{ACD}. Not accounting for the continual change in many cybersecurity tasks diminishes the potential for, and evaluation of, long term agent performance. 
        
        \textbf{Security Implications of D-NS} In practice, adversaries, network conditions, and software systems evolve continuously; yet agents trained under stationary assumptions learn on fixed dynamics and predictable behavior. Such agents are brittle to change and exploitable in dynamic environments, where they can fail by adopting obsolete strategies, or be manipulated by adversarial tactics. Beyond adversarial evolution, broader concept drift (\eg firewall or software updates) can degrade performance over time, causing silent failures undermining long-term performance. Critically, systems may appear effective under static evaluation providing a false sense of robustness, but perform unreliably, once deployed in the ever-changing landscape of cybersecurity operations.

        \textbf{Recommendations for D-NS} Cybersecurity tasks are inherently dynamic, so DRL agents must be explicitly exposed to such variability during training. Environments should be designed for adaptation; for example, non-stationarity can be incorporated through temporal shifts, randomized parameters, evolving topologies, and adaptive adversaries \eg~\cite{zhao2021structural}. Additionally, techniques such as curriculum learning, domain randomization, and population-based training can further help agents generalize to progressively hardening conditions.
        Furthermore, adversarial training should be used where appropriate, having both sides evolve iteratively (alternating or concurrently updating attacker and defender policies), \eg~\cite{cortellazzi2024train,goel2024optimizing}. Training agents against adaptive counterparts simulates the attacker-defender arms race, enabling agents to adapt~\cite{tsingenopoulos2023adversarial}. Brittleness, and long-term resilience can then be measured through transfer tests against unseen strategies. The intentional embedding of non-stationarity in environments means \ac{DRL} agents understand the evolving security landscape.

    \subsection{Case Study of Deployment Pitfalls}\label{sec:DeployingCaseStudy}

        We evaluate the impact of pitfalls in deploying \ac{DRL} agents using MiniCAGE~\cite{emerson_cyborg_2024} (introduced in Sections~\ref{sec:case-study-envs} and~\ref{sec:TrainingCaseStudy}).
        First, we consider \uafull by modifying the assumed fixed action order between attacker and defender during testing, blue then red (B$\rightarrow$R) in the original formulation. Second, we study \nsfull by changing the fixed attacker strategy between training and testing. 

        \textbf{Breakdown of Underlying Assumption} The first section of Table~\ref{tab:deploy_CS_table} shows the average reward of the \ac{DRL} defender when varying the action order of red and blue agents in training and testing (\eg R$\rightarrow$B indicates red then blue and \emph{Mixed} indicates random turn order).
        As expected, agents perform comparably when tested on the same training order (-17.2 for B$\rightarrow$R and -15.8 for R$\rightarrow$B). When the assumed order changes, performance degrades: minimally for R$\rightarrow$B agent (-21.0) and drastically for the B$\rightarrow$R agent (-72.7). 
        These results highlight how minor design choices can degrade performance if the assumptions they rely on do not hold in practice. 

        \input{Tables/Deploy-CS-Table}
    
        \textbf{Changing Attacker Strategy} Table~\ref{tab:deploy_CS_table} shows the average reward of defender agents under varying attacker strategies: B-Line and Meander (see Section~\ref{sec:case-study-envs}). The agent order is kept constant as B$\rightarrow$R. 
        We see a degradation in performance when any stationary blue agent is evaluated against unseen or mixed attacker strategies. 
        This degradation can be mitigated by including non-stationary attacker strategies during training as the Mixed blue agent outperforms the stationary defenders in both mixed and cross evaluations. 
        These results show that policies can be brittle to changing conditions when trained without accounting for non-stationarity.

\section{Limitations} \label{sec:limitations}
By selecting a representative set of academic works, we aim to show that the pitfalls identified are present even in highly-cited or top conference papers. Thus, while extending our analysis to the entirety of \drlsec literature would improve the coverage of pitfall prevalence, it would provide limited insight beyond the findings of this work. To minimize bias in reviewing we (a) were lenient in cases of ambiguity, (b) employed a two-reviewer process with 81.5\% initial agreement, and (c) resolved disagreements through discussion. To highlight pitfalls we performed didactic case studies, these are intended as illustrative examples where pitfalls have measurable, concrete, consequences in security environments. The specific consequence of a pitfall depends on the security problem, how it is modeled, implemented, algorithm chosen, and numerous other factors. We encourage researchers to carefully consider the pitfalls of \ac{DRL} for cybersecurity and, where applicable, apply the recommendations from this work.

\section{Related Work} \label{sec:related-work}

To the best of our knowledge, ours is the first work to delineate the pitfalls that arise when applying \ac{DRL} in security.
Many papers, such as Sommer and Paxson~\cite{sommer2010outside} discuss use of ML in \emph{different} cybersecurity tasks; we focus our discussion on works identifying the pitfalls in applying ML \emph{paradigms} across cybersecurity.
In 2022, Arp~\etal~\cite{DosDonts} identified 10 pitfalls in 30 supervised learning-based security research papers, finding at least three pitfalls in each paper. More recently, Evertz~\etal~\cite{evertz2025chasing} extended this analysis to LLM-based security research. They identified nine pitfalls, finding at least one in each of the 72 papers reviewed. While there are some parallels between previously identified pitfalls and our work, the sequential decision-making paradigm of \ac{DRL} is different. This means that although similar in spirit, these parallel ideas manifest in fundamentally different ways. For example, invalid environment modeling opposed to incorrect dataset construction: they have different causes, impacts, and mitigations. Complementing these pitfall-focused papers, several papers have surveyed the application of DRL to specific cybersecurity tasks~\cite{nguyen2021deep,vyas2025towards,mavroudis2025guidelines,hicks2026building}, or security of DRL~\cite{lei_new_2023}.
Although these studies provide insights in their respective paradigm and application areas, there remains a gap in systematizing the pitfalls specific to \ac{DRL} for cybersecurity. DRL introduces fundamentally different challenges: from the sequential nature of decision-making to the reliance on environment design~\cite{bates2023reward,bates2025less}. 
Therefore, by identifying the major shortcomings in current \drlsec literature and providing actionable recommendations, we lay the groundwork for future high-quality research.

\section{Conclusion} \label{sec:conclusion}
The autonomous and adaptive decision-making that DRL can offer in complex environments has significant potential for cybersecurity applications. However, our review of 66 \drlsec papers published between 2018-2025 reveals 11 common methodological challenges (pitfalls) that undermine \drlsec research. Most commonly, we find that: 71.2\% lack clear evidence of policy convergence, 60.6\% fail to address partial observability, 66.7\% neglect variance analysis, and 40.9\% evaluate in oversimplified environments. Through case studies in ACD, adversarial malware creation, and web security testing, we demonstrate these pitfalls directly impact performance and create potentially brittle policies. Through the identification of 11 \drlsec pitfalls and recommendations for their mitigation, this work establishes higher methodological standards for future applications of \ac{DRL} to cybersecurity.

\section*{Acknowledgments}
This research was partially funded and supported by: 
the UK National Cyber Security Centre (NCSC);
UK EPSRC Grant no. EP/X015971/2;
the Defence Science and Technology Laboratory (DSTL), an executive agency of the UK Ministry of Defence, supporting the Autonomous Resilient Cyber Defence (ARCD) project within the DSTL Cyber Defence Enhancement programme; a Google Academic Research Award (GARA); the Research Fund KU Leuven; the Cybersecurity Research Program Flanders.

\appendix

\section*{Ethical Considerations} \label{app:ethics}
 
This work identifies common methodological pitfalls in published research. We emphasize that these pitfalls represent \emph{systematic} challenges in adapting DRL to cybersecurity rather than failures of individual researchers. Our goal is constructive: to provide concrete, actionable, recommendations that strengthen future research. We frame our analysis around pitfall patterns over critiquing specific papers, and while we list the papers reviewed, \emph{we do not report the per-paper pitfalls}.

\textbf{Fairness in review} Our review examined 66 papers published between 2018 and 2025. To ensure fairness, each paper was independently evaluated by two reviewers using predefined pitfall definitions, with disagreements resolved through discussion and the benefit of the doubt given when pitfall presence was ambiguous (see Section~\ref{sec:review-method}). Further breakdown of agreement, kappa, and interpretation can be found in Appendix~\ref{app:keywords}.

\textbf{Stakeholders and potential harms} 
Our paper could theoretically affect several stakeholders.
First, this work sets higher standards for \textit{researchers}, however, may be perceived as criticism; we mitigate this by reporting only aggregate prevalence and by framing pitfalls as community-wide problems. 
Second, by explicitly describing the problems in \drlsec literature and by providing recommendations we aim to inform \textit{practitioners} to mitigate these failure modes in practice. 
Third, our discussion could inform \textit{adversaries} to potential weaknesses in DRL-based defenses; however, these vulnerabilities are already implicit in existing publications and, therefore, improving \drlsec research provides more value to the security community than it does to potential attackers. 
Finally, other stakeholders, including organizations that depend on DRL-based tools, receive indirect gains from improvements in \drlsec research.

\textbf{Ethics Decision} We conclude that this research is ethical because: (1) it provides substantial benefit to the security research community by identifying and addressing systematic methodological challenges; (2) it does not expose individuals to harm or violate their rights; (3) it treats analyzed prior work fairly and constructively; and (4) the potential for misuse is minimal compared to the benefits of improving research methodology. The decision to publish serves the public interest by advancing the quality and reliability of DRL-based cybersecurity research.

\section*{Open Science} \label{app:open}

To support open science and promote reproducibility, we provide the following repositories.
\textbf{GitHub}: \url{https://github.com/alan-turing-institute/DRL4Sec-Pitfalls}
\newline \textbf{Zenodo}: \url{https://doi.org/10.5281/zenodo.20209122}

\textbf{Case Study Environments} We provide the version of the environment used to create the results in this paper. This includes: training and evaluation scripts alongside any changes made to the environment to ablate the impact of pitfalls. The complete list of environments is: MiniCAGE ACD environment (Sections~\ref{sec:TrainingCaseStudy}, \ref{sec:DeployingCaseStudy}), Link XSS environment ( Section~\ref{ModelingCaseStudy}), \sqirl\ SQL Injection environment (Section~\ref{sec:EvaluatingCaseStudy}), and the AutoRobust adversarial ML environment (Section~\ref{sec:EvaluatingCaseStudy}). Alongside the code we also report the MDPs and hyperparameters for each environment in repositories.

\textbf{Literature Review Data} The following information from our literature review can be found as follows: (1) a complete list of the 66 papers analyzed with bibliographic information (reported through corpus characteristics, Section~\ref{sec:review-method}, and associated references); (2) the search keywords (reported in our repositories); and, (3) the inclusion/exclusion criteria used (reported in Section~\ref{sec:review-method} and our repositories).
We have chosen to not release the per-paper pitfalls, similar to ~\cite{DosDonts,evertz2025chasing}, in order to keep the constructive nature of the paper.

\bibliographystyle{plain}
\bibliography{ref}

\appendix

\begin{figure*}[h]
    \centering
    \includegraphics[width=1\textwidth]{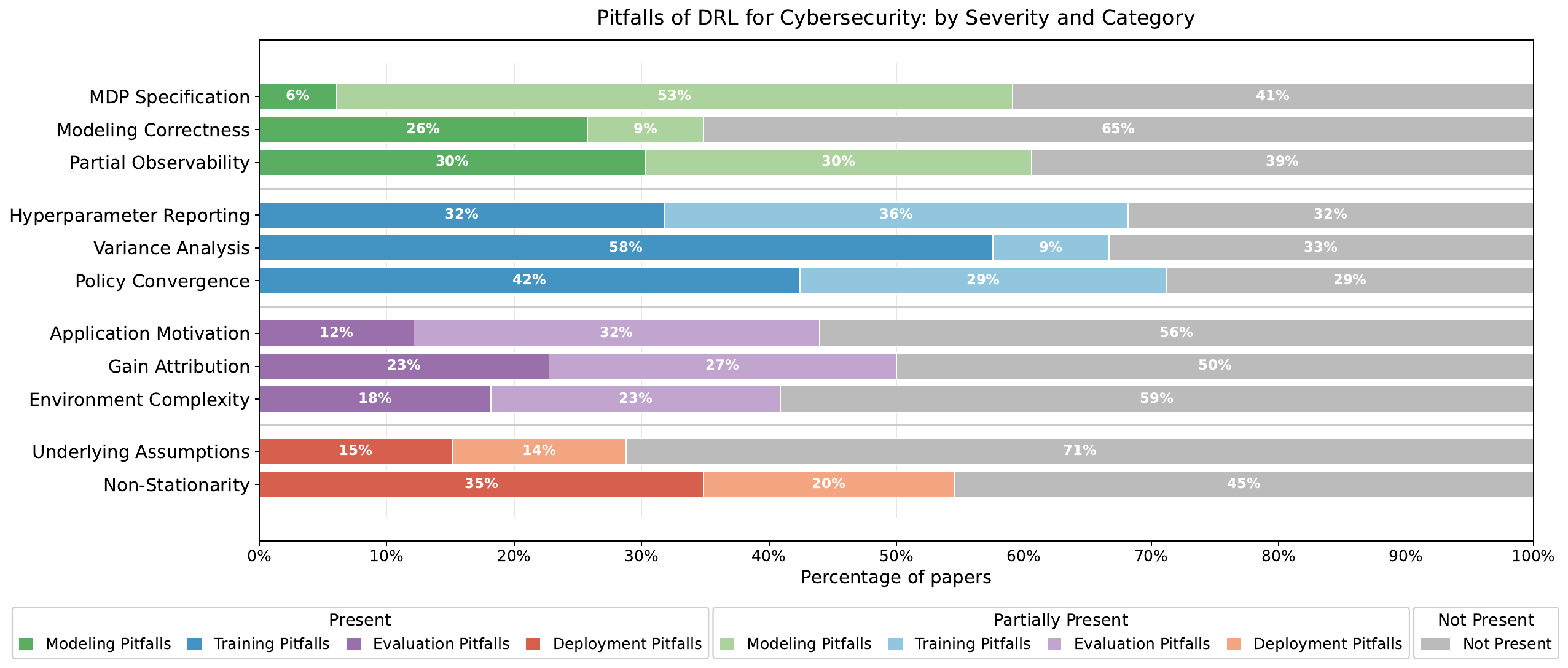}
    \caption{Complete breakdown of pitfall prevalence}
    \label{fig:pitfallComplete}
\end{figure*}

\section{Search \& Review} \label{app:keywords}

\textbf{Keywords} We defined the following keywords after reviewing the proceedings of top security conferences (\eg USENIX Security, IEEE S\&P, ACM CCS, and NDSS) in order to collect a representative set of research on DRL being applied to differing cybersecurity tasks. ("Deep Reinforcement Learning") \textit{and} (cybersecurity \textit{or} "web security" \textit{or} fuzzing \textit{or} "vulnerability discovery" \textit{or} "vulnerability detection" \textit{or} "network defence" OR "network defense" \textit{or} malware \textit{or} evasion \textit{or} sqli \textit{or} xss \textit{or} "penetration testing" \textit{or} "cyber defence" \textit{or} "cyber defense" \textit{or} "network security" \textit{or} Dos \textit{or} "intrusion detection"). 

\textbf{Included Publishers} Outside of explicitly included A*/A Venues: IEEE, ACM, Elsevier, Springer Nature, and PMLR

\textbf{Corpus Venue Breakdown} The breakdown of papers per venue is as follows. \textit{A*/A Conferences (24 papers)}: USENIX Security (3),  AAAI (3),  RAID (3), CCS (2), Euro S\&P (2), ISSTA (2), ACSAC (2), ASE (2), NDSS (1), Web Conf (1), NeurIPS (1), HPCA (1), ECCV (1). \textit{Journals (31 Papers)}: IEEE Transactions (7), Elsevier (7), Springer Nature (5), ESORICS (5), IEEE Journals (3), Computers\&Security (2), ACM Transactions (2). \textit{Other Venues (11)}: IEEE Conferences (6), Miscellaneous (5). 

\textbf{Review} The breakdown of reviewer agreement by pitfall can be found in Table~\ref{tab:review-agreement} and the prevalence across all pitfalls can be found in Figure~\ref{fig:pitfallComplete}.

\begin{table}[h]
        \caption{Breakdown of the reviewer agreement across pitfalls.}
        \label{tab:review-agreement}
        \vspace{1ex}
        \centering
        \begin{adjustbox}{width=\linewidth}
        \begin{tabular}{llll}
        \toprule
        \multicolumn{1}{c}{\textbf{Pitfall}} & \multicolumn{1}{c}{\textbf{Agreement}}  & \multicolumn{1}{c}{\textbf{Kappa}} & \multicolumn{1}{c}{\textbf{Interpretation}}  \\ 
        \midrule
            \multirow{1}{*}{\ms} & 75.8\% & 0.579 & Moderate  \\
                            \addlinespace
            \multirow{1}{*}{\mc} & 89.4\% & 0.797 & Substantial  \\
                            \addlinespace
            \multirow{1}{*}{\po} & 72.7\% & 0.583 & Moderate  \\
                            \addlinespace
            \multirow{1}{*}{\hr} & 86.4\% & 0.796 & Substantial \\
                            \addlinespace
            \multirow{1}{*}{\va} &  89.4\%  & 0.798 & Substantial \\
                            \addlinespace
            \multirow{1}{*}{\pc} &  81.8\%  & 0.720 & Substantial \\
                            \addlinespace
            \multirow{1}{*}{\am} &  87.9\%  & 0.789 & Substantial \\
                            \addlinespace
            \multirow{1}{*}{\ga} &  74.2\%  & 0.594 & Moderate \\
                            \addlinespace
            \multirow{1}{*}{\ec} &  80.3\%  & 0.654 & Substantial \\
                            \addlinespace
            \multirow{1}{*}{\ua} &  80.3\%  & 0.570 & Moderate \\
                            \addlinespace
            \multirow{1}{*}{\ns} &  78.8\%  & 0.663 & Substantial \\
                            \addlinespace
        \bottomrule
        \end{tabular}
        \end{adjustbox}
    \end{table}

\section{Case Study Environments} \label{app:case_studies}
\subsection{MiniCAGE} \label{app:minicage_mdp}

\textbf{State space} The blue agent state space consists of:
A 78 value vector containing information about each host. The first 52 values relate to the red action and level of compromise, split into groups of 4 values per host. The first two values indicate if any relevant red activity has occurred, and the next two indicate the level of red compromise:
\medskip

\noindent
\begin{minipage}[t]{0.48\linewidth}
\centering
\textbf{Red activity}\\[0.5em]
\begin{tabular}{ll}
\textbf{Values} & \textbf{Meaning} \\
\hline
0, 0 & None \\
1, 0 & Scan \\
1, 1 & Exploit \\
\end{tabular}
\end{minipage}
\hfill
\begin{minipage}[t]{0.48\linewidth}
\centering
\textbf{Red access level}\\[0.5em]
\begin{tabular}{ll}
\textbf{Values} & \textbf{Meaning} \\
\hline
0, 0 & No access \\
1, 0 & Unknown access \\
0, 1 & User level \\
1, 1 & Privilege level \\
\end{tabular}
\end{minipage}
\vspace{1ex}

The next part of the observation space contains scan information. This is an additional 13 float values, one for each host. The value can be: 2 if a the host is being scanned in the current timestep, 1 if the host was scanned in a prior timestep, and otherwise 0.
The final part of the observation space is for tracking the decoys placed. This is also represented as an additional vector of 13 float values for each host, each value indicating the number of available decoys per host remaining to be placed.

\textbf{Action space} The blue agent action space consists of 53 possible actions and includes the global action `sleep', as well as host specific actions:
\begin{enumerate*}[label={(\arabic*)}]
    \item Analyze: Reveals further information about the given host to better allow blue to identify if the red agent is present. I.e Identifying if privileged access has been achieved.
    \item Decoy: Places a decoy service on the given host. There are a set available per host.
    \item Remove: Attempt to remove red access from a host, given privileged access has not be achieved. This could remove red access after a successful red attempt to exploit network services, but not after a successful red escalate action.
    \item Restore: Restores a host to its initial known good state. Whilst this removes all red access, it has consequences for system availability.
\end{enumerate*}

The Red action space consists of 56 possible actions, also including the global `sleep' and the Discover Remote systems action that acts upon each subnet: User, Enterprise and Operational. There are host specific actions:
\begin{enumerate*}[label={(\arabic*)}]
    \item Discover Network Services: This discovers responsive services on the given host.
    \item Exploit Network Services: Attempts to exploit a specific service on the remote system.
    \item Escalate: Escalates the agent's privilege level on the given host. If this succeeds, the host is fully compromised.
    \item Impact: This action can only be applied to the Operational Server. It disrupts the performance of the network and repeating this action is the final goal for the red agents.
\end{enumerate*}

\textbf{Reward Function} The reward function consists of a series of penalties based upon the level of red access on hosts, red actions and blue actions.
\begin{enumerate*}[label={(\arabic*)}]
    \item -0.1: Per user host compromised
    \item -1.0: Per enterprise host compromised
    \item -1.0: Per operational server compromised
    \item -0.1: Per operational host compromised
    \item -10: Each successful red impact action
    \item -1.0: Each time blue restores a host
\end{enumerate*}

\textbf{Transition Function} Transition dynamics are implicitly defined by the MiniCAGE network environment. At each timestep, Red and Blue actions are selected from the same pre-step state (Red first, then Blue). The global state evolves via rule-based host-level dynamics which decide environment aspects like red compromise progression, service configuration, and Blue defending actions. Stochasticity arises mainly from partial observability, though there is some deliberately induced noise (i.e there is a 5\% chance that a red exploit action is not reflected in the immediate blue observation space).

\textbf{Model architecture} StableBaselines3's PPO was implemented using the default MLP actor-critic architecture. This includes a shared feature extractor, policy and value heads each with 2 x 64 fully connected layers and ReLU activation functions.

\textbf{Hyperparameters} Aside from when these are explicitly changed in~\ref{sec:TrainingCaseStudy}, the hyperparameters used throughout the MiniCAGE experiment can be seen in Table~\ref{tab:hyperparameters_cage}.
\begin{table}[h]
\centering
\caption{MiniCAGE Training hyperparameters.}
\vspace{1ex}
\label{tab:hyperparameters_cage}
\begin{tabular}{ll}
\hline
\textbf{Hyperparameter} & \textbf{Value} \\
\hline
Learning rate & $3 \times 10^{-4}$ \\
Discount factor ($\gamma$) & 0.99 \\
Clip Range ($\epsilon$) & 0.2 \\
Epochs & 6 \\
Batch size & 64 \\
n steps & 2048 \\
Lambda ($\lambda$) & 0.95 \\
Entropy coefficient & 0 \\
Value loss coefficient & 0.5 \\
\hline
\end{tabular}
\end{table}

\subsection{Link}
\label{app:link-mdp}

\textbf{State Space}
The state space of the Link environment consists of a vector of 47 integers that represent:
\begin{enumerate*}[label={(\arabic*)}]
    \item The payload input, including information such as the payload appearance and the payload repetitiveness
    \item The HTML page response, which encodes: the reflected payload appearance, and payload context information.
\end{enumerate*}

\textbf{Action Space} Each of the 39 actions of the Link environment can be placed into one of 8 categories: 
\begin{enumerate*}[label={(\arabic*)}]
    \item Basic Payload (4 actions): Designed to initialize a payload with a simple payload \eg\ \code{<script>alert(1);</script>}
    \item JS Component (3 actions): generate a JavaScript snippet \eg\ \code{alert(1);}
    \item Prefix (13 actions): Add symbols to the front of the payload \eg\ \code{<, ', `}
    \item Suffix (3 actions): Add symbols to the end of the payload \eg \code{<!--}
    \item Tag (3 actions): Actions to manipulate the tag: \eg\ capitalize the tag.
    \item Attribute (1 action): Capitalize the payload attribute.
    \item JS snippet (5 actions): Manipulate the JavaScript snippet \eg\ converting \code{alert(1)} to  \code{throw(1)}
    \item Entire String (6 actions): Manipulate the entire payload \eg\ obfuscate by octet encoding.
\end{enumerate*}

\textbf{Reward Function}
The reward function of this environment first reward if a vulnerability is found, reward for the difference between the maximum and current number of steps. However, a penalty of -1 is applied: at each timestep or when the payload doesn't change between steps.
Finally, the frequency of the current payload divided by the maximum training steps is subtracted at each step.

\textbf{Transition Function}
At each timestep the agent selects an action to perform, as described above. These are applied to the payload in a rule based manor: actions are added to the payload or substituted when an action artifact is already in the payload, 
\eg replacing \code{img} HTML tag with \code{video} HTML tag. The action for media tags (img, video, audio, svg),  event attributes (onerror, onload, onclick, onmouseover), and  JS snippets (alert(1), confirm(1), prompt(1)) insert at random from a predefined set. 
Termination of the episode occurs when a vulnerability has been found or $t >= 500$.

\textbf{Model Architecture} Following from Lee \etal~\cite{lee2022link} we use a an actor critic architecture from StableBaselines3 of three layers of 128 neurons. 

\textbf{Hyperparameters} $\alpha = 5 \times 10^{-4}$, $\gamma = 0.95$, timesteps $3.5 \times 10^6$.

\subsection{\sqirl} \label{app:SQiRL_mdp}

\textbf{State Space} A state space vector of 2049 features, 1024 features from the Gated Recurrent Unit (GRU) representation of the payload, 1024 from a GRU representation of the SQL statement executed, and a single value that represents an error occurring from the injected payload. 

\textbf{Action Space} The action space of \sqirl\ has a base set of 27 tokens to add to the payload. However it is dynamic as these can be added or removed at any location in the payload. Token actions can be split into three categories: 1) Basic Tokens (\eg commas, comments, quote marks), 2) Behavior Changing tokens which include SQL keywords (\eg OR, AND, IF), 3) Sanitization Escape including obfuscation techniques such as capitalization, whitespace and SQL keyword encoding.

\textbf{Reward Function} \sqirl\ uses two rewards, an internal reward based on Random Network Distillation (RND), reward for finding new states~\cite{burda2018exploration}. It also uses an external reward that penalizes -1 at each timestep a vulnerability is found, and 0 when it is found. 

\textbf{Transition Function} Deterministic $T: \mathcal{S} \times \mathcal{A} \rightarrow \mathcal{S}$. Termination: on vulnerability found or $t > 30$.

\textbf{Model Architecture} Following from Al Wahaibi \etal~\cite{al2023sqirl} we use their pretrained embedding models for SQLi payloads and SQL statements, and a DQN architecture of 3 layers, 2048, 1024, 512 nodes per layer.

\textbf{Hyperparameters} $\alpha = 5\times 10^{-3}$, $\gamma = 0.99$, rollout $N=1024$, batch $B=512$, episodes per vulnerability $200$.

\subsection{AutoRobust} \label{app:autorobust_mdp}

\textbf{State Space} 
$\mathcal{S} \subset \mathbb{R}^{768}$: [CLS] embeddings of a fine-tuned DistilBERT. For reports $>512$ tokens, mean-pooled chunk embeddings with 20-token overlap.

\textbf{Action Space} 
Multi-discrete $\mathcal{A} = \{0,1\} \times \{0,\ldots,12\} \times \{0,1,2,3\}$: $(a_{\text{op}}, a_{\text{key}}, a_{\text{gen}})$ where $a_{\text{op}}$ selects between adding a goodware entry or editing an existing malware entry, $a_{\text{key}}$ selects the dynamic analysis report category to operate on, and $a_{\text{gen}}$ specifies the entry replacement strategy (dictionary word, token from goodware corpus, random alphanumeric, or random choice between the previous 3).

\textbf{Reward Functions}
$r_t^{(1)} = p_{t-1} - p_t$,
and
$r_t^{(2)} = p_{t-1} - p_t + \frac{100}{t}$ if $p_t < 0.5$, otherwise $r_t^{(2)} = p_{t-1} - p_t$,
where $p_t$ is probability of malware at step $t$.

\textbf{Transition Function}
Deterministic $T: \mathcal{S} \times \mathcal{A} \rightarrow \mathcal{S}$. Given a modified report $R_t \rightarrow R_{t+1}$, the next state is $s_{t+1} = \phi(R_{t+1})$ with $\phi: \mathcal{R} \rightarrow \mathbb{R}^{768}$ the [CLS] embedding. Termination: $p_t < 0.5$ or $t \geq 1000$.

\textbf{Model Architecture}
PPO with MLP policy: $[768 \rightarrow 128 \rightarrow 128 \rightarrow |\mathcal{A}|]$, ReLU activations.

\textbf{Hyperparameters}
$\alpha = 3 \times 10^{-3}$, $\gamma = 0.99$, rollout $N=1024$, batch $B=32$, timesteps $5 \times 10^4$.

\end{document}

%% file: Tables/Link-mini-table.tex
\begin{table}[t]
\caption{Percent of vulnerabilities found (VF\%) and 95\% CI results for Link over 20 runs demonstrating the impact of modeling choices.}
\label{tab:modelingCaseStudy}
\vspace{1ex}
\centering
\begin{tabular}{lcc}
\toprule
\multicolumn{1}{c}{\multirow{2}{*}{\textbf{MDP}}}   & \multicolumn{2}{c}{\textbf{Vulnerabilities Found}}  \\ 
\cline{2-3}
& \multicolumn{1}{c}{\text{\small VF (\%)}} & \multicolumn{1}{c}{\text{\small 95\% CI}} \\
\midrule \addlinespace
\multirow{1}{*}{Original}       
                & 75.2  & (59.4, 90.9) \\
                \addlinespace
\multirow{1}{*}{Distinct States}  
                    & 85.3 & (76.7, 94.0) \\ 
                    \addlinespace
\bottomrule
\end{tabular}
\end{table}

%% file: Tables/SQiRL_EC-CS-Table.tex
\begin{table}[t]
\caption{Percent of vulnerabilities found (VF\%) and 95\% CI for SQiRL across 20 runs and evaluations each setting. Showing the impact of evaluating on low-complexity environments. 
}\label{tab:sqirl}
\vspace{1ex}
\begin{adjustbox}{width=0.98\linewidth}
\begin{tabular}{lllll}
\toprule
\multirow{3}{*}{%
  \parbox{1.2cm}{\centering \vspace{6pt} \textbf{Training\\Environment}}
} & \multicolumn{4}{c}{\textbf{Evaluation Environment}} \\ \cmidrule{2-5} 
 & \multicolumn{2}{c}{Sanitized} & \multicolumn{2}{c}{Non-Sanitized} \\
 \cmidrule(lr){2-3}
 \cmidrule(lr){4-5}
 & \multicolumn{1}{c}{\small{VF (\%)}} & \multicolumn{1}{c}{\small{95\% CI}} & \multicolumn{1}{c}{\small{VF (\%)}} & \multicolumn{1}{c}{\small{95\% CI}} \\ \midrule
Combined & 41.10 & (33.9, 48.4) & 88.5 & (85.4, 91.6)  \\
Non-Sanitized & 29.80 & (22.9, 36.8) &  88.7 & (85.5, 91.9) \\ \bottomrule
\end{tabular}
\end{adjustbox}
\end{table}

%% file: Tables/Deploy-CS-Table.tex
\begin{table}[t]
\centering
\caption{Experimental results for MiniCAGE deploying agents case study. The top section shows blue agent performance when training and evaluation action orders differ. The bottom section shows blue agent performance against different red adversaries. Zero is the theoretical max score.}
\vspace{1ex}
\begin{adjustbox}{width=0.98\linewidth}
\begin{tabular}{@{}p{1.1cm}cccccc@{}}
\toprule
\multirow{3}{*}{%
  \parbox{1.2cm}{\centering \vspace{6pt} \textbf{Training\\Order}}
} &
  \multicolumn{6}{c}{\textbf{Evaluation Order}} \\
\cmidrule(l){2-7}
& \multicolumn{2}{c}{\textbf{R$\rightarrow$B}} 
& \multicolumn{2}{c}{\textbf{B$\rightarrow$R}} 
& \multicolumn{2}{c}{\textbf{Mixed}} \\
\cmidrule(lr){2-3} \cmidrule(lr){4-5} \cmidrule(lr){6-7}
& \text{\small Score} & \text{\small 95\% CI}
& \text{\small Score} & \text{\small 95\% CI}
& \text{\small Score} & \text{\small 95\% CI} \\
\midrule
R$\rightarrow$B &\textbf{ -15.8} & (-14.8, -16.7) & -21.0 & (-19.0, -23.1) & \textbf{-18.3} & (-17.0, -19.6) \\
B$\rightarrow$R & -72.7 & (-65.6, -79.9) & \textbf{-17.2} & (-16.7, -17.7) & -45.5 & (-42.0, -49.0) \\ 
Mixed           & -28.4 & (-25.6, -31.2) & -19.9 & (-19.2, -20.6) & -24.6 & (-22.9, -26.2) \\ 

\midrule
\multirow{3}{*}{%
  \parbox{1.2cm}{\centering \vspace{6pt} \textbf{Training\\Red\\Agent}}
} &
  \multicolumn{6}{c}{\textbf{Evaluation Adversary}} \\
\cmidrule(l){2-7}
& \multicolumn{2}{c}{\textbf{B-line}} 
& \multicolumn{2}{c}{\textbf{Meander}} 
& \multicolumn{2}{c}{\textbf{Mixed}} \\
\cmidrule(lr){2-3} \cmidrule(lr){4-5} \cmidrule(lr){6-7}
& \text{\small Score} & \text{\small 95\% CI}
& \text{\small Score} & \text{\small 95\% CI}
& \text{\small Score} & \text{\small 95\% CI} \\
\midrule
B-line  & \textbf{-19.0} & (-20.9, -17.2) & -104.1 & (-120.7, -87.6) & -61.4 & (-70.5, -52.3) \\
Meander & -79.0 & (-97.2, -60.9) & \textbf{-61.0}  & (-74.7, -47.3) & -70.2 & (-83.5, -56.9) \\
Mixed   & -40.2 & (-47.1, -33.4) & -79.2  & (-86.5, -71.9) & \textbf{-60.2} & (-65.8, -54.7) \\
\bottomrule
\end{tabular}
\end{adjustbox}
\label{tab:deploy_CS_table}
\end{table}